\documentclass[runningheads]{llncs}

\usepackage{eccv}

\usepackage{eccvabbrv}

\usepackage{graphicx}
\usepackage{booktabs}
\usepackage{amsmath,amssymb,amsfonts}
\usepackage{booktabs}
\usepackage{multirow}
\usepackage{xcolor}
\usepackage{hyperref}
\usepackage{makecell}
\usepackage{xcolor}
\usepackage{gensymb}
\usepackage{rotating}

\usepackage[accsupp]{axessibility}  

\hypersetup{colorlinks=true,citecolor=eccvblue,breaklinks=true}

\usepackage{orcidlink}

\begin{document}

\title{iMED: A Multi-Endoscope Dataset for Surgical 3D Perception} 

\titlerunning{iMED}


\title{iMED: A Multi-Endoscope Dataset for Surgical 3D Perception}
\titlerunning{iMED}

\author{
Sierra Bonilla\inst{1} \and
Fengyi Jiang\inst{2} \and
Chinedu Nwoye\inst{2} \and
Jingpei Lu\inst{2} \and
Kailey Reardon\inst{2} \and
Humphrey W. Chow\inst{2} \and
Francisco Vasconcelos\inst{1} \and
Sophia Bano\inst{1} \and
Adam Schmidt\inst{2} \and
Omid Mohareri\inst{2}
}

\authorrunning{S. Bonilla et al.}

\institute{
University College London, London WC1E 6BT, United Kingdom \and
Advanced Research, Intuitive, Sunnyvale, CA 94086, USA \\
\email{sierra.bonilla.21@ucl.ac.uk}
}

\maketitle

\begin{abstract}
We introduce iMED, the first synchronized multi-endoscope dataset for robot-assisted minimally invasive surgery (RAMIS), designed to address a fundamental limitation in surgical vision benchmarking: the absence of independent held-out viewpoints. Existing surgical datasets capture single-trajectory sequences from a single endoscope, making it impossible to distinguish geometric generalization from photometric interpolation along a narrow forward-facing path. iMED provides 340 sequences ($\approx$170K synchronized timepoints with 4 views per timepoint) recorded simultaneously from two independent stereo endoscopes across ex vivo, postmortem, and live surgical settings, spanning 14 specimens with diverse anatomical regions and motion regimes. The dataset includes calibrated camera intrinsics, ArUco-based frame-wise pose estimates with uncertainty quantification, instrument segmentation masks, and rich clinical metadata. 
Using a train-on-one-endoscope, test-on-another evaluation protocol, we benchmark 23 state-of-the-art methods across rigid and deformable novel view synthesis, pose estimation, feature matching, and monocular depth estimation. Our experiments show that methods relying primarily on photometric supervision degrade substantially under this held-out-endoscope setting, while methods with explicit geometric regularization are more robust across views in our benchmark.
Notably, large-scale foundation models transfer surprisingly well to surgical imagery, often outperforming domain-specific models, suggesting pitfalls in current fine-tuning protocols. iMED establishes a new evaluation protocol for geometric generalization in surgical vision. Dataloader and dataset links can be found at
\href{https://github.com/surgical-vision/imed}{\texttt{github.com/surgical-vision/imed}}.


\keywords{Surgical vision \and Robot-assisted minimally invasive surgery \and 3D from multi-view}   
\end{abstract}

\section{Introduction}
\label{sec:introduction}

\begin{figure}[t]
    \centering
    \includegraphics[width=0.83\columnwidth]{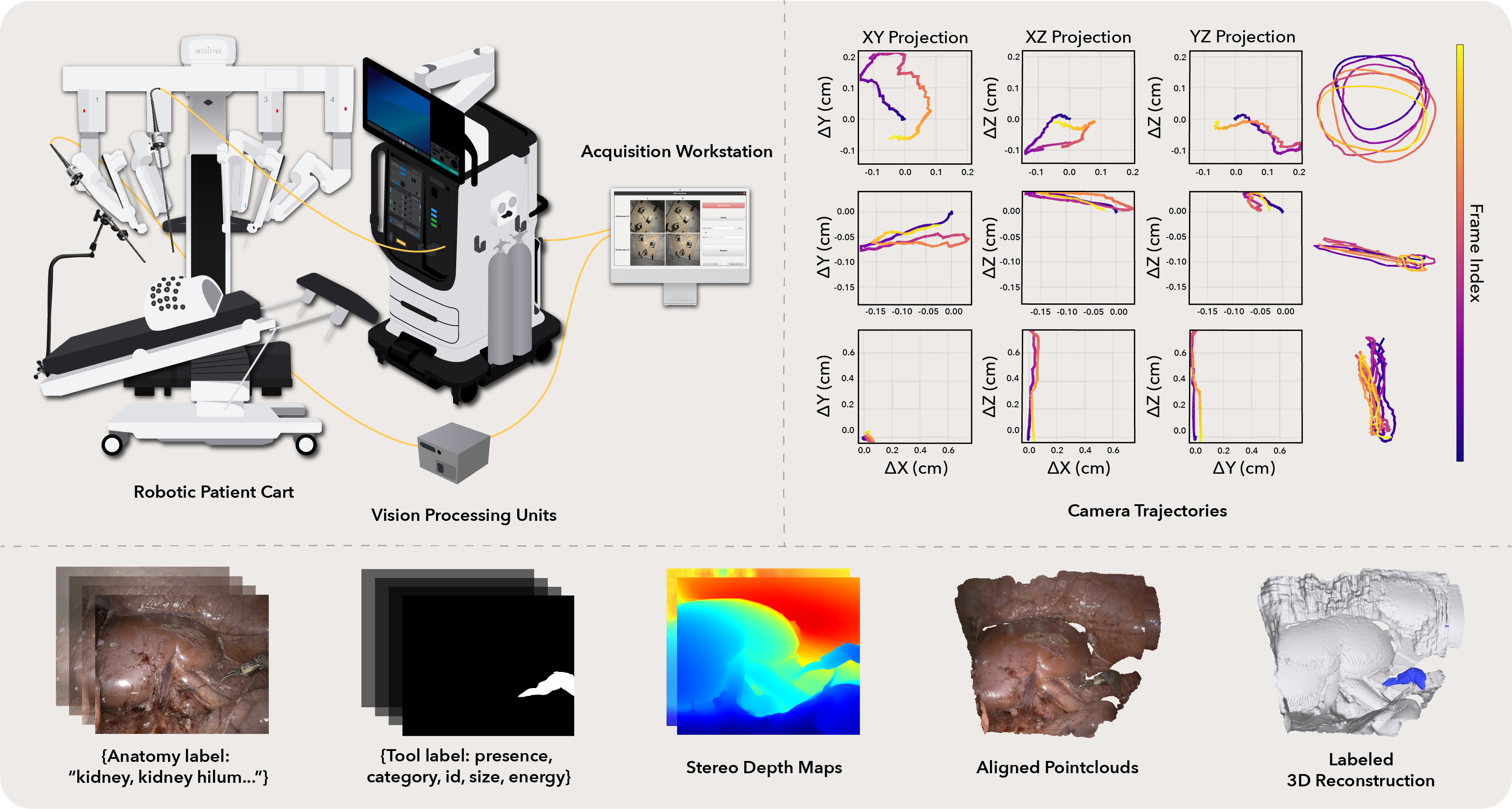}
    \caption{Data platform with two stereo endoscopes (top-left), common camera trajectories (top-right), and pipeline outputs including session/anatomy/instrument/motion labels, camera intrinsics, ArUco pose estimates, inferred tool masks , stereo depth (code provided), and derived 3D geometry. See~\cref{tab:label_provenance} for provenance.}
    \label{fig:teaser1}
\end{figure} 

Non-rigid scenes with constrained camera motion are among the most challenging regimes for image-based 3D perception tasks (i.e. pose estimation, feature matching, optical flow, depth estimation, and novel view synthesis (NVS)). Although challenging, such conditions often arise in real-world robotic applications like digital human modeling of clothed bodies~\cite{bertiche2020cloth3d}, submarine exploration~\cite{chadebecq2020refractive}, and robot-assisted minimally invasive surgery (RAMIS)~\cite{schmidt2024surgical}. Among these, RAMIS represents an extreme regime. Cameras operate inside the body under severe viewpoint constraints, typically following narrow, forward-facing trajectories through deformable tissue with strong specularities and transient phenomena (smoke, bleeding, topology changes). These conditions compress multiple failure modes for 3D perception methods, making RAMIS an informative stress test for visual geometry in real-world robotic systems.

Computer vision benchmarks have historically advanced 3D perception through standardized evaluation protocols~\cite{geiger2012we}. In rigid environments, datasets capturing scenes from many surrounding viewpoints with known camera poses enabled progress in multi-view reconstruction and mapping~\cite{aanaes2016large,knapitsch2017tanks,dai2017scannet,yeshwanth2023scannet++}. Dynamic-scene benchmarks later extended these ideas to time-varying NVS and reconstruction~\cite{pumarola2021d,park2021nerfies,park2021hypernerf,Joo_2015_ICCV}, typically relying on wide-baseline or 360$^\circ$/180$^\circ$ capture where cameras observe scenes from diverse directions. Surgical vision datasets have enabled progress in reconstruction, camera tracking, and tool analysis~\cite{allan2021stereo,schmidt2024surgical,hayoz2023pose,hartwig2022miti}. Unlike the surrounding-view capture in general benchmarks, endoscopic cameras follow narrow forward-facing paths through tissue, producing sequences with significant image overlap and limited viewpoint diversity.

\begin{table*}[!t]
    \centering
    \caption{Comparison of stereo-endoscopic datasets for 3D perception in minimally invasive surgery. Multi-Camera* defined as more than one stereo camera pair. **only for 2 phantom validation sequences with CT scans.}
    \label{tab:dataset_comparison}
    \setlength\tabcolsep{0.01in}
    \resizebox{0.998\linewidth}{!}{
    \begin{tabular}{@{}lcccccrcccrccrcc@{}}
    \toprule
    \multirow{2}{*}{Dataset}  & \multicolumn{5}{c}{Dataset Attributes} & \phantom{abc} & \multicolumn{3}{c}{Scene Types} & \phantom{abc} & \multicolumn{2}{c}{Inferred Labels} & \phantom{abc} & \multicolumn{2}{c}{Provided Reference Signals} \\
    \cmidrule{2-6} \cmidrule{8-10} \cmidrule{12-13} \cmidrule{15-16}
    
    & Seq. & Specimens & Modality & Setting & Live && Static & \makecell{Dyn. Rigid} & \makecell{Deform. Obj.} && \makecell{3D Model} & \makecell{Tool Masks} && \makecell{Camera Calib.} & \makecell{Multi-Camera*} \\
    \midrule

    \makecell[l]{EndoVis 17/18\\~\cite{allan20192017, allan20202018}} & 29 & 29 & \makecell{Da Vinci Xi\\Da Vinci X} & \textit{In vivo} & \checkmark && &\checkmark&\checkmark && & \checkmark && & \\
    MITI~\cite{hartwig2022miti} & - & 1 & \makecell{Karl Storz\\3D Tipcam} & \textit{In vivo} & \checkmark && & & \checkmark && & && \checkmark & \\
    STIR~\cite{schmidt2024surgical} & 572 & 4 & Da Vinci Xi & \makecell{\textit{Ex vivo}\\\textit{In vivo}} & \checkmark && & & \checkmark && & && \checkmark & \\
    
    \makecell[l]{Hamlyn Centre\\~~\cite{mountney2010three, stoyanov2010real, pratt2010dynamic}} & - & - & Da Vinci Si & \makecell{Phantom\\\textit{Ex vivo}\\\textit{In vivo}} & \checkmark && \checkmark & & \checkmark && \checkmark** & && & \\
    SERV-CT~\cite{edwards2022serv} & 16 & 2 & Da Vinci & Postmortem & && \checkmark & & && \checkmark & && \checkmark & \\
    SCARED~\cite{allan2021stereo} & 9 & 9 & Da Vinci Xi & Postmortem &  && \checkmark & & && \checkmark & && \checkmark & \\
    StereoMIS~\cite{hayoz2023pose} & 16 & 6 & Da Vinci Xi & \textit{In vivo} & \checkmark && & \checkmark & \checkmark && & \checkmark && \checkmark & \\
    \midrule
    \textbf{iMED (Ours)} & 340 & 14 & Da Vinci 5 & \makecell{\textit{Ex vivo}\\\textit{In vivo}\\Postmortem} & \checkmark && \checkmark & \checkmark & \checkmark && \checkmark & \checkmark && \checkmark & \checkmark \\
    
    \bottomrule
    \end{tabular}
    }
\end{table*}

In NVS and 3D scene representation, models are typically evaluated using images at held-out camera poses and reporting metrics such as PSNR and SSIM against held-out views~\cite{barron2022mip,seidenschwarz2025dynomo,hayoz2024online,yang2024deformable,kerbl20233d}. These metrics quantify photometric fidelity and have become standard quantitative benchmarks. However, they are proxies for reconstruction quality rather than direct measures of geometric consistency: high PSNR/SSIM can reflect local interpolation between nearby viewpoints, without uniquely determining whether the model has learned the underlying 3D structure~\cite{wickrema2025benchmarking}. This limitation is amplified when train and test poses lie along the same forward-facing trajectory, where test views are typically sampled from nearby frames in time so differ only slightly in view. A complementary evaluation is to test generalization under larger viewpoint shifts. In classic multi-camera dynamic-scene benchmarks, this protocol is well established: models are trained on a subset of cameras and evaluated on held-out cameras~\cite{Joo_2015_ICCV,broxton2020immersive,li2022neural}.

To address this limitation in surgical vision, we introduce \textbf{iMED}, the first multi-endoscope dataset for RAMIS, providing synchronized recordings from two physically independent stereo endoscopes, four camera streams with overlapping fields of view, enabling cross-trajectory geometric validation (see~\cref{fig:teaser1}). The dataset comprises 340 sequences totaling $\sim$170K synchronized timepoints (4 frames$\times$60 FPS) across diverse surgical scenarios: 10 \textit{ex vivo} organ specimens (chicken, porcine, and bovine; $\approx$7 sequences each), 3 human cadaver subjects ($\approx$6 anatomical sites per subject, $\approx$7 sequences per site), and 1 live porcine subject (10 anatomical sites, $\approx$7 sequences per site). Data collection was performed in a USDA-licensed clinical lab using one porcine model and three human cadaver models sourced from a willed body program. Each 10-15 second sequence captures varied conditions including static scenes, respiratory motion, tool-tissue interactions with 20 common robotic instruments (see both~\cref{tab:label_provenance} and Suppl.Tab.7, contact-based tissue deformations, lighting variations, and specularities. All sequences include comprehensive metadata: anatomical regions labeled by clinical specialists with surgical robotics expertise, robot tool types, powered instrument usage (e.g., electrocautery), and motion categories (e.g., camera sweep, robot tool-tissue interaction, calibration). All frames include inferred filtered instrument segmentation masks generated by~\cite{carion2025sam}. Refer to~\cref{tab:label_provenance} for label provenance. To filter sequences towards specific applications (e.g. deformable vs. rigid), refer to Suppl.Tab.10. Recordings include intrinsic calibrations and frame-wise ArUco-based~\cite{garrido2014automatic} camera poses refined through reprojection-error optimization with uncertainty estimates.

\begin{figure}[t]
    \centering

    \begin{minipage}[t]{0.53\columnwidth}
        \centering
        \includegraphics[width=\linewidth]{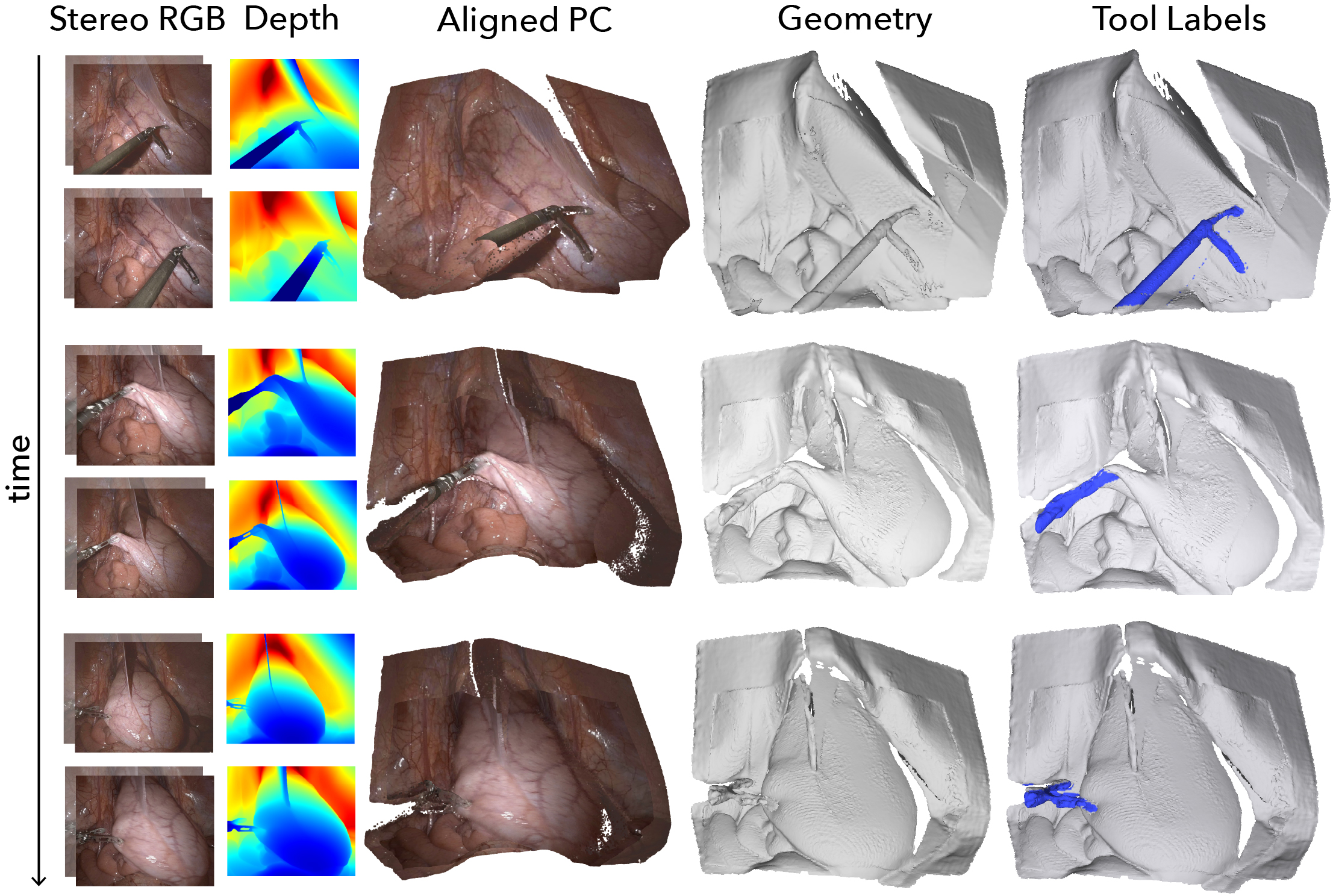}
    \end{minipage}
    \hfill
    \begin{minipage}[t]{0.33\columnwidth}
        \centering
        \includegraphics[width=\linewidth]{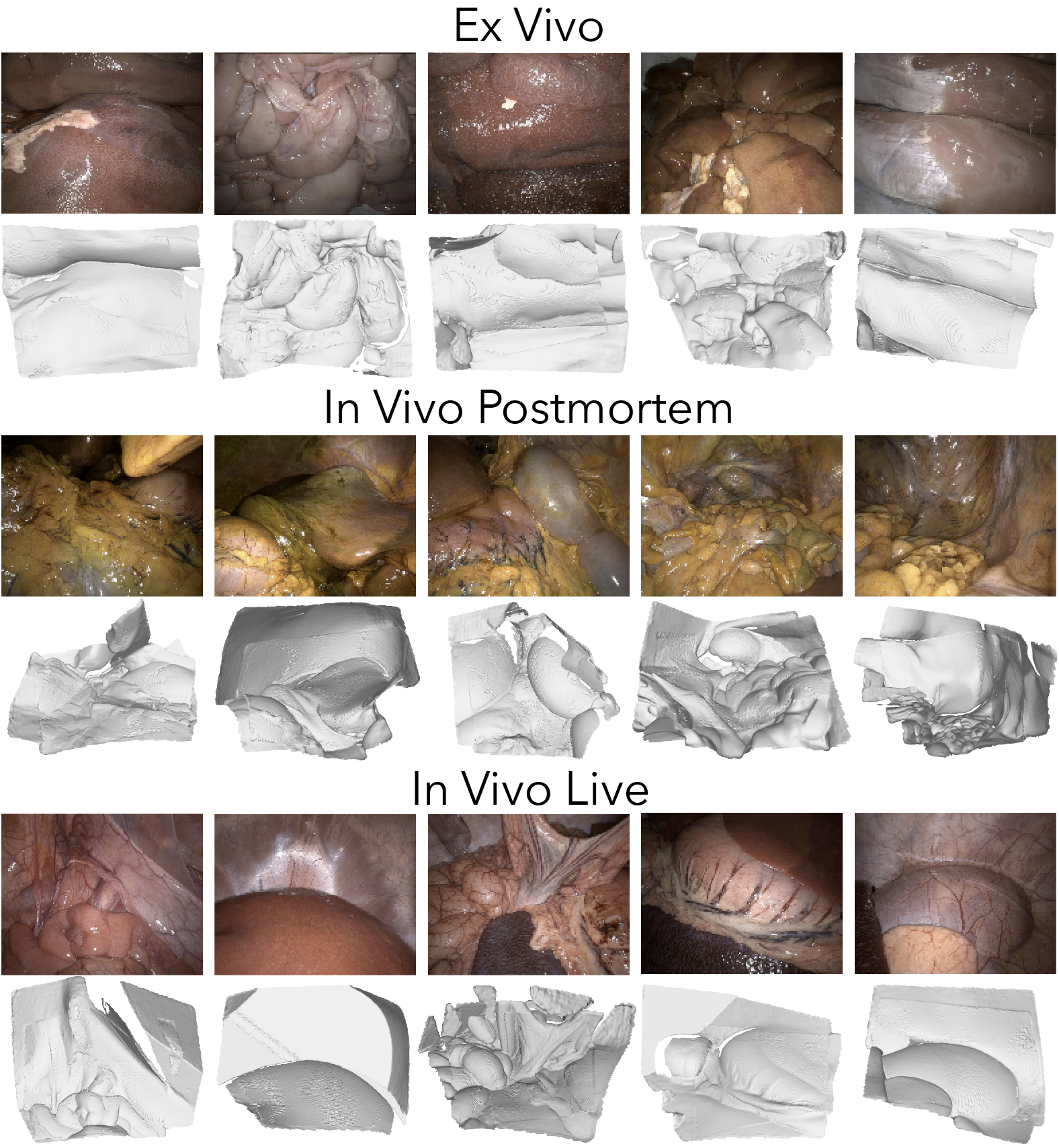}
    \end{minipage}

    \caption{
    Example geometric reconstructions and label projection enabled by iMED.
    \textit{Left:} synchronized stereo views from both endoscopes, stereo depth estimationo~\cite{wen2025foundationstereo}, aligned point clouds from provided camera calibrations, Poisson-reconstruction, and reprojected tool masks.
    \textit{Right:} Representative anatomical regions with surface reconstructions from \textit{ex vivo} specimens (chicken, porcine, bovine), human cadaver sites, and live porcine tissue.
    Inference code to recreate geometry and 3D tool labels is provided.
    }
    \label{fig:teaser2}
\end{figure}

\begin{figure}[t]
    \centering
    \includegraphics[width=0.93\columnwidth]{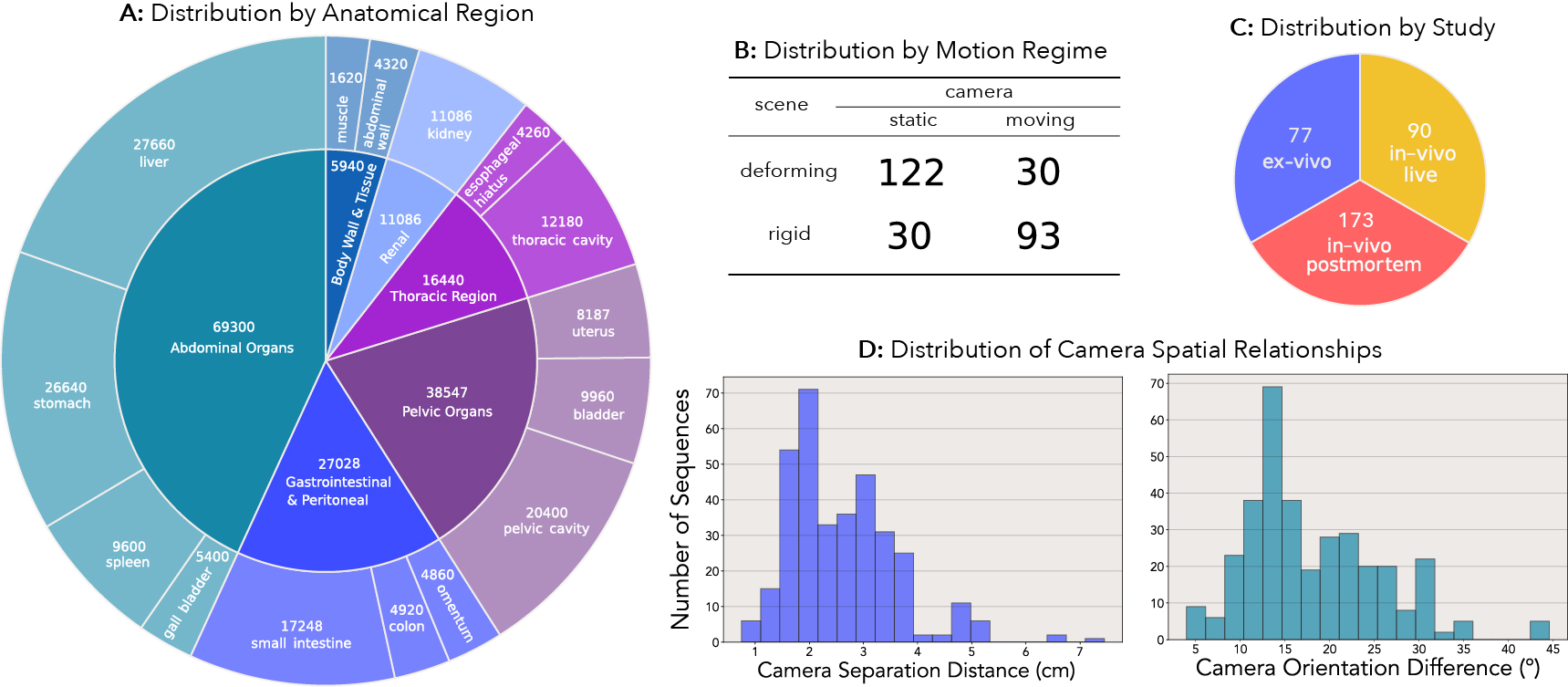}  
    \caption{
    \textit{A:} Distribution of seqs. by anatomical region and organ type including number of timepoints per category. 
    \textit{B:} Distribution of seqs. by motion regime (camera motion vs. scene motion). Calibration-only sequences are excluded. 
    \textit{C:} Distribution of seqs. by session type. 
    \textit{D:} Distribution of seqs. by cam spatial relationships, showing inter-cam separation distance and orientation difference.
    }
    \label{fig:stats}
\end{figure}

Using iMED, we evaluate 23 SOTA algorithms across tasks central to 3D perception: rigid NVS, deformable (online \& not) NVS, camera pose estimation, feature matching, and depth estimation. 
Our analysis suggests that, under this held-out-endoscope protocol, methods relying primarily on photometric supervision degrade under physically independent viewpoints, while methods with explicit geometric regularization are more robust across views. Depth supervision has representation-dependent effects, and large multi-domain foundation models often transfer well to surgical imagery compared with surgical-only training or aggressive fine-tuning. We define train-on-one-endoscope, test-on-another evaluation protocols that eliminate trajectory overlap and measure geometric generalization across physically independent cameras, a standard in multi-view human capture datasets that is absent from surgical benchmarks. Motivated by the lack of independent test views in surgical datasets, we introduce iMED and make the following contributions:
\begin{enumerate}
    \item A train-on-one-endoscope, test-on-another evaluation protocol providing held-out viewpoints with large baseline and viewing-direction differences, as a better proxy for geometric generalization beyond photometric interpolation
    \item A synchronized multi-endoscope dataset for RAMIS with 340 sequences ($\approx$170K timepoints with 4 images per timepoint) across \textit{ex vivo, in vivo} postmortem, and \textit{in vivo} settings (See~\cref{fig:stats})
    \item Calibrated multi-view images and structured annotations (cam intrinsics, reference trajectories, anatomical labels, motion categories, instrument masks)
    \item A benchmark of 23 methods across NVS, pose estimation, feature matching, and monocular depth estimation that exposes previously unobserved behaviors: models trained using only image supervision perform poorly on new viewpoints, methods with explicit 3D regularization generalize more reliably, and foundation models often transfer better than surgical-only training
\end{enumerate}  
\section{Related Work}
\label{sec:related}

\textit{Multi-View and 3D Reconstruction Datasets.} Rigorous evaluation of 3D reconstruction requires validated geometric ground truth. Multi-view stereo benchmarks establish ground truth through controlled conditions: DTU~\cite{aanaes2016large} employs robotic arms capturing 49-64 viewpoints with trajectories; BlendedMVS~\cite{yao2020blendedmvs} synthesizes depth maps from 20-1000 input views per scene via photogrammetric reconstruction; Tanks and Temples~\cite{knapitsch2017tanks} validates outdoor reconstructions from 150-300+ images against laser scans. These methods enable pixel-accurate depth evaluation but are confined to static, well-textured scenes. Object-centric datasets (CO3D, uCO3D~\cite{liu2025uncommon, reizenstein2021common}) capture objects from smartphone video, while unbounded scene datasets like Mip-NeRF 360~\cite{barron2022mip} and RealEstate10K~\cite{zhou2018stereo} use COLMAP~\cite{schoenberger2016sfm} and ORB-SLAM2~\cite{mur2017orb} on controlled captures. In many of these datasets, static scenes are recorded from a wide range of viewpoints, often circling around the object or environment. Because the same scene is observed from many directions, it is logical to train models on some views and evaluate them on other viewpoints from the same capture.

Dynamic scenes introduce an additional challenge: views captured at different moments no longer correspond to the same geometry. Several datasets study deformable reconstruction under these conditions. D-NeRF~\cite{pumarola2021d} and Nerfies~\cite{park2021nerfies} explore neural rendering from monocular video, while HyperNeRF~\cite{park2021hypernerf} uses a stereo rig. These works typically focus on predicting intermediate time steps, training on subsets of frames and testing on unseen frames along the same trajectory. Multi-view human capture datasets instead measure spatial generalization directly by recording scenes with many synchronized cameras. PanopticStudio~\cite{Joo_2015_ICCV, Joo_2017_TPAMI} (480 cameras), Immersive Video Dataset~\cite{broxton2020immersive} (46 cameras), and Plenoptic Video~\cite{li2022neural} (21 cameras) train models on some viewpoints and evaluate them on others captured at the same moment in time. Because the held-out views come from different cameras, this setup provides a stronger test of the geometry than interpolation along a single trajectory. These datasets assume isolated deforming subjects against largely static backgrounds, a configuration incompatible with surgical scenes where the entire field of view may deform. The Drunkard’s Dataset~\cite{recasens2023drunkard} further studies deformable visual odometry under exploratory camera motion using synthetic data designed with endoscopic navigation in mind, where the entire scene undergoes non-rigid distortion.

\begin{table}[t]
\centering
\caption{Label provenance for iMED. We report annotation level, generation type, source method, and whether the annotation is included in the public data release. For storage reasons, stereo depth and 3D geometry are reproducible via provided code.}
\label{tab:label_provenance}
\scriptsize
\setlength{\tabcolsep}{3pt}
\begin{tabular*}{\textwidth}{@{\extracolsep{\fill}}llllc}
\toprule
Label & Level & Provenance & Source / Method & Released \\
\midrule
Session type & Seq. & Manual & Curator & \checkmark \\
Broad anatomical region & Seq. & Manual & Clinical specialist & \checkmark \\
Detailed scene caption & Scene & Manual & Clinical specialist & \checkmark \\
Instrument type & Seq. & Manual & Clinical specialist & \checkmark \\
Motion regime/movement type & Seq. & Manual & Curator + protocol & \checkmark \\
Camera intrinsics/stereo baseline & Seq. & Measured & Calibration procedure & \checkmark \\
Frame-wise camera poses & Frame & Measured & ArUco PnP + refinement & \checkmark \\
Instrument masks & Frame & Inferred & ~\cite{carion2025sam} & \checkmark \\
ArUco masks & Frame & Inferred & ~\cite{carion2025sam} & \checkmark \\
ArUco in-painted & Frame & Inferred & ~\cite{li2025diffueraser} & \checkmark \\
Stereo depth & Frame & Inferred & ~\cite{wen2025foundationstereo} (code provided) & $\times$ \\
3D geometry (point clouds/meshes) & Frame & Derived & Poisson (code provided) & $\times$ \\
\bottomrule
\end{tabular*}
\end{table}
\textit{Surgical Datasets.} Surgical datasets face unique ground truth challenges. The physical impossibility of placing dozens of cameras inside the body constrains surgical vision systems to far fewer synchronized viewpoints than multi-view benchmarks. Early surgical datasets established ground truth through external sensing modalities: SCARED~\cite{allan2021stereo} uses structured light across 9 sequences; SERV-CT~\cite{edwards2022serv} aligns endoscopic views with CT scans but provides only 16 static image pairs from two specimens; EndoAbS~\cite{penza2018endoabs} employs laser scanning for 120 image pairs across 20 fixed phantom poses. These datasets, captured from postmortem tissue or phantoms, enable stereo depth evaluation but are fundamentally limited to single-trajectory, single-timepoint capture in non-deforming environments. Recent datasets expanded temporal and deformation coverage while maintaining single-trajectory capture: STIR~\cite{schmidt2024surgical} provides 572 sequences with infrared marker tracking; StereoMIS~\cite{hayoz2023pose} contributes 16 sequences with poses; MITI~\cite{hartwig2022miti} features stereo with calibrations within one procedure; SegSTRONG-C~\cite{ding2024segstrong} targets tool segmentation providing binary instrument masks over a small set of sequences (17) \textit{ex vivo} animal tissues. Despite increased sequence counts, all existing surgical datasets provide single endoscope captures along a single path through time. iMED fills these gaps (real-world deformation, multi-camera capture) for surgical settings through synchronized dual-endoscope capture (4 camera streams) with overlapping fields of view, enabling cross-trajectory spatial validation across 340 sequences spanning \textit{ex vivo}, postmortem, and live surgical scenes (examples of anatomical regions in ~\cref{fig:teaser2}) with ArUco-based reference pose estimation.
\section{Dataset Curation}
\label{sec:dataset_curation}


\subsection{Acquisition Set-up}

Data was acquired using three Anonymized surgical systems with three 8mm 0° and four 8mm 30° endoscopes. Due to hardware constraints, only one endoscope mounts to the robotic arm operationally. For \textit{in vivo} sessions, the second endoscope was placed on an articulated passive arm (See~\cref{fig:teaser1}, top left); for \textit{ex vivo} sessions, both endoscopes were hand-guided in non-operative configuration. Throughout this paper, we designate the stationary endoscope mounted on the articulated arm as \textit{Endoscope 1}, which remains fixed during data collection. The mobile endoscope mounted on the robotic patient cart with active illumination is designated as \textit{Endoscope 2}. Only \textit{Endoscope 2}'s integrated light source was activated during acquisition to maintain lighting conditions representative of standard MIS procedures. Since the two endoscopes observe different illumination characteristics, we normalize their photometric appearance using sequence-level statistics to ensure robustness against per-frame artifacts. See Suppl. Sec. 6.4 for evaluation of multiple normalization strategies.

We capture four HDMI streams from two hardware-synchronized endoscopes separated by 2.6±1.2 cm with 18.7°±10.0° orientation difference (\cref{fig:stats}). This 2.6cm inter-camera separation is 
substantially larger than the 4mm intra-endoscope stereo baseline, 
representing up to 50\% of the typical 6-13 cm laparoscopic field of 
view~\cite{wang2017endoscope,supe2010ergonomics}. Synchronization accuracy across $\approx$170K frame pairs achieved mean timing difference of 3.36 ms (SD = 3.90 ms), much less than the $\approx$17ms frame duration. See Suppl. Sec. 6.1 for complete hardware synchronization methodology and timing analysis. All sequences are compressed using H.264 CRF 18 encoding; we recommend GPU-accelerated decoding (NVDEC) for efficient processing.

\begin{figure}[t]
\centering

\begin{minipage}{0.48\columnwidth}
    \centering
    \includegraphics[width=\linewidth]{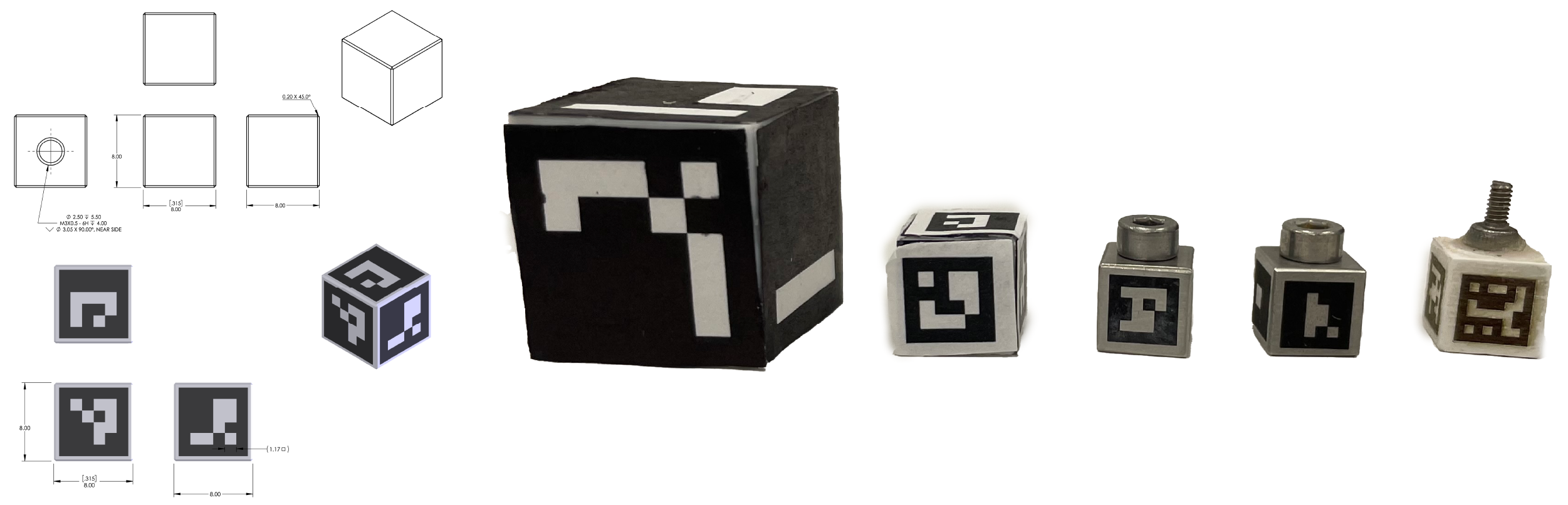}
\end{minipage}
\hfill
\begin{minipage}{0.47\columnwidth}
    \centering
    \includegraphics[width=\linewidth]{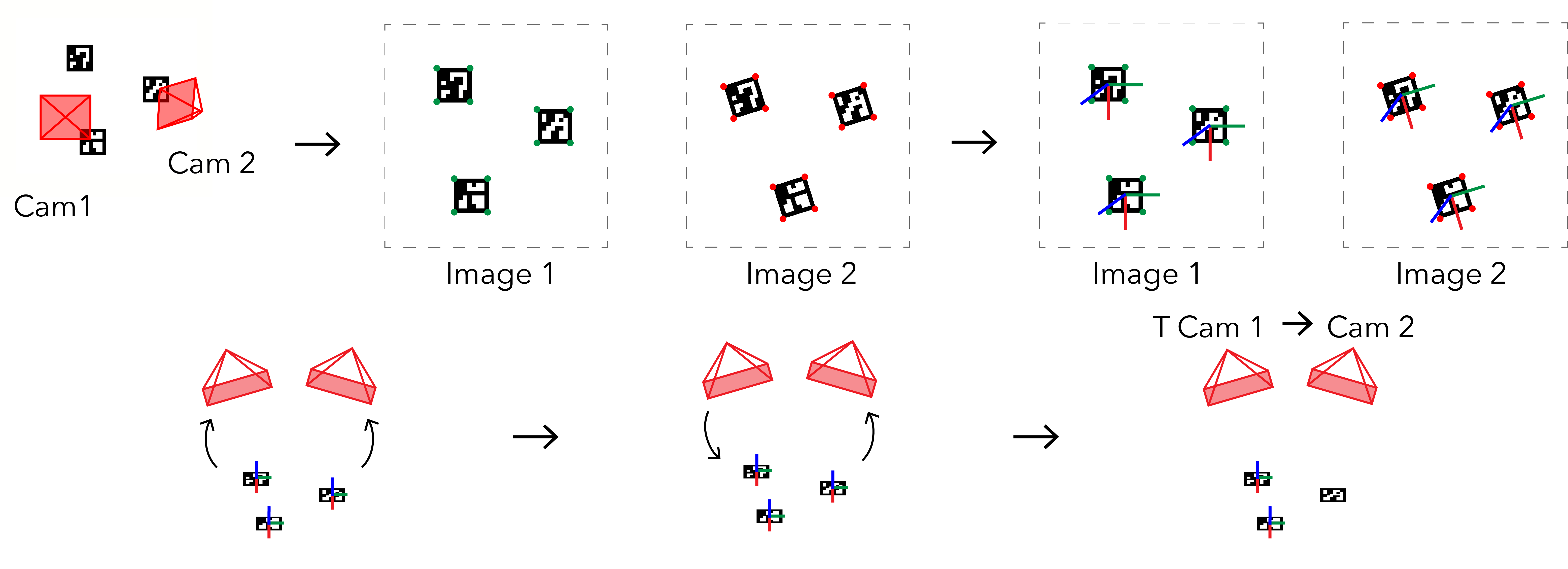}
\end{minipage}

\caption{
\textit{Left:} ArUco calibration cube design iterations leading to our final 8\,mm wooden cube design. 
\textit{Right:} ArUco-based pose estimation pipeline: (1) marker detection, (2) PnP pose estimation, 
(3) relative transformation via RANSAC and refinement.
}
\label{fig:aruco_combined}
\end{figure}


\subsubsection{Calibration Strategy}
Due to hardware constraints preventing direct capture of relative pose between the two endoscopes, we developed an external calibration system using custom calibration cubes with ArUco fiducial markers~\cite{garrido2014automatic}. ArUco markers are square fiducial tags with an internal bit pattern designed for pose estimation and unique identification. Each marker consists of an outer perimeter of black cells surrounding an inner data payload region (examples seen in~\cref{fig:aruco_combined}). Our design requirements included: (1) fit through standard trocar ports ($\le$12mm~\cite{limperg2021laparoscopic}), (2) provide high-contrast markers for reliable detection, (3) be manipulable by robotic instruments, (4) remain stable via surface tension when placed on tissue, and (5) resist staining during acquisition sessions. After multiple design iterations (~\cref{fig:aruco_combined}), we converged on 8mm wooden cubes with 5.7mm laser-etched ArUco markers from the "\texttt{ARUCO\_MIP\_36h12}" dictionary~\cite{garrido2016generation}. See Suppl. Sec. 6.1 for complete design iteration analysis.

We use two acquisition protocols, filterable via metadata flags: (1) \textit{Static Camera Post-Calibration} ($\approx$40\% of sequences): ArUco cubes are moved around to densely sample the working volume, then removed before tissue manipulation sequences proceeds with fixed cameras (mean drift pre and post tissue manipulation: 1.97° rot, 3.3mm trans, indicating stable calibration throughout manipulation tasks); (2) \textit{Dynamic Camera Sequences} ($\approx$40\% of seqs): markers remain visible throughout to enable continuous pose estimation during movement. 

\subsection{Reference Camera Poses}
\label{subsec:ground_truth_poses}


To provide the reference camera trajectories, we employ an ArUco marker-based pose estimation pipeline computing relative transformations between endoscopes (\cref{fig:aruco_combined}). For each 5.7mm ArUco marker detected in both cameras, we estimate 6-DOF poses using a Perspective-n-Point solver~\cite{itseez2015opencv} with known marker geometry and camera intrinsics. We aggregate observations across a sliding window and compute rel. camera-to-camera transformations for commonly visible markers. 

The full pipeline consists of five stages: (1) \textit{RANSAC Initialization and Refinement:} To handle detection errors and occlusions, we use RANSAC~\cite{Fischler1981} to identify inlier observations based on angular and translational consistency thresholds. We then refine the best hypothesis by minimizing bidirectional reprojection error across all inlier marker corners using Levenberg-Marquardt optimization~\cite{levenberg1944method,marquardt1963algorithm}. (2) \textit{Stereo Fusion:} Each endoscope contains left and right stereo cameras. We independently compute left-to-left and right-to-right transformations, then average via quaternion interpolation for the final trajectory. For static calibration sequences, the sliding window spans the entire sequence duration, providing a globally-optimized transformation from all observations. (3) \textit{Outlier Detection:} We identify outliers through frame-to-frame velocity analysis. Expert laparoscopic tool velocities are less than 8.5 cm/s~\cite{hwang2006correlating}. At 60 fps capture rate, this corresponds to frame-to-frame displacements of approximately 0.14 cm. Considering this and a working volume of 20cmx20cmx20cm, we set conservative thresholds rejecting poses with translation changes ${>}0.7$ cm/frame and rotation changes ${>}3.0\degree$/frame to effectively filter ArUco detection errors and mechanical artifacts while preserving plausible motion. (4) \textit{Temporal Smoothing:} We apply Savitzky-Golay filtering  ($W=15$ frames, $p=1$) to translation axes and smooth rotations in quaternion space with sign continuity enforcement and renormalization. Outlier poses use linear interpolation (translation) and SLERP~\cite{shoemake1985animating} (rotation) between smoothed neighbors. (5) \textit{Uncertainty Quantification:} We assess pose reliability with consistency between independently computed left-to-left and right-to-right transformations, quantifying translational (in m) and rotational (in $^\circ$) uncertainty. Ablation studies validate each stage of the trajectory pipeline, showing up to 90\% reprojection err. reduction and sub-2\,mm physical accuracy at 5\,cm working distance. We release trajectories, raw calibration seqs, and recomp code; summary results appear in \cref{tab:trajectory_summary}, (ablations in Suppl~6.5).



\begin{table}[t]
\centering
\caption{Reference trajectory accuracy after full refinement across session types. See Supplementary Sec.~6.5 for complete ablations and 3D consistency metrics.}
\label{tab:trajectory_summary}
\scriptsize
\setlength{\tabcolsep}{6pt}
\begin{tabular}{lcc}
\toprule
Session Type & Reproj. Err. (px)$\downarrow$ & Physical Err. @5cm (mm)$\downarrow$ \\
\midrule
Ex vivo & 38.33 & 1.88 \\
In vivo (postmortem) & 17.00 & 0.83 \\
In vivo (live) & 11.87 & 0.58 \\
\bottomrule
\end{tabular}
\end{table}
\section{Benchmarking Experiments}
\label{sec:benchmarking}

We evaluated 23 SOTA methods: two rigid NVS, six deformable NVS, 11 feature matching/pose estimation methods, and four monocular depth estimation methods. Perhaps not surprisingly, methods designed for wide baseline multi-view datasets struggle when confronted with surgical constraints. The train-on-endoscope-2, test-on-endoscope-1 protocol exposes a failure mode: methods that interpolate photometric appearance collapse when forced to extrapolate geometric structure from constrained viewpoints. Performance correlates with geometric regularization, but at the cost of computation time, a challenge for real-time constrained 3D perception. All experiments were run on four NVIDIA RTX 6000 Ada GPUs. For reproducibility, the sequences are labeled (Suppl. Tab. 10). For the NVS, pose, and feature-matching tasks, reference cam-to-cam transforms are estimated from calibration observations in the static-camera protocol; dynamic-camera trajectories are released as a resource and can be filtered using metadata.

\begin{table}[t]
\centering
\caption{NVS benchmarks. Methods trained on \textit{Endoscope 2} and evaluated on \textit{Endoscope 1}. Rigid and deformable methods perform offline optimization vs. online methods incrementally update the scene using only the most recent observations during acquisition (see~\cref{results:rigid}). FPS excludes SfM~\cite{schoenberger2016sfm} preprocessing.}
\label{tab:nvs-benchmark}
\scriptsize
\setlength{\tabcolsep}{3pt}
\begin{tabular*}{\linewidth}{@{\extracolsep{\fill}}lccccc@{}}
\toprule
Method & Depth & PSNR$\uparrow$ & SSIM$\uparrow$ & \makecell{Render (FPS)$\uparrow$} & \makecell{Train (FPS)$\uparrow$ }\\
\midrule
\multicolumn{6}{c}{\textbf{Rigid Novel View Synthesis}} \\
\midrule
NeRF\cite{mildenhall2021nerf} & x & 17.24 & 0.65 & 0.11 & 0.21 \\
NeRF\cite{mildenhall2021nerf} & DA\cite{yang2024depth} & 16.29 & 0.55 & 0.11 & 0.19 \\
NeRF\cite{mildenhall2021nerf} & FS\cite{wen2025foundationstereo} & 15.93 & 0.55 & 0.11 & 0.18 \\
3DGS\cite{kerbl20233d} & x & 14.82 & 0.59 & 487.62 & 0.59 \\
3DGS\cite{kerbl20233d}  & DA\cite{yang2024depth} & 14.77 & 0.63 & 509.99 & 0.36 \\
3DGS\cite{kerbl20233d}  & FS\cite{wen2025foundationstereo} & 15.33 & 0.63 & 497.43 & 0.31 \\
\midrule
\multicolumn{6}{c}{\textbf{Deformable Novel View Synthesis}} \\
\midrule
4DGS\cite{yang2023real} & x & 13.74 & 0.57 & 578.46 & 0.38 \\
D-3DGS\cite{yang2024deformable} & x & 14.50 & 0.51 & 31.41 & 0.14 \\
4DGS\cite{wu20244d} & x & 15.65 & 0.33 & 38.99 & 0.76 \\
Endo-4DGS\cite{huang2024endo} & DA\cite{yang2024depth} & 16.26 & 0.59 & 309.37 & 0.66 \\
\midrule
\multicolumn{6}{c}{\textbf{Online Deformable Novel View Synthesis}} \\
\midrule
Hayoz\cite{hayoz2024online} & RAFT\cite{teed2020raft} & 15.66 & 0.40 & 224.15 & 0.15 \\
Hayoz\cite{hayoz2024online} & FS\cite{wen2025foundationstereo} & 15.92 & 0.40 & 229.39 & 0.14 \\
DynOmo\cite{seidenschwarz2025dynomo} & DA\cite{yang2024depth} & 24.16 & 0.87 & 65.18 & 0.07 \\
DynOmo\cite{seidenschwarz2025dynomo} & FS\cite{wen2025foundationstereo} & 25.29 & 0.90 & 63.58 & 0.06 \\
\bottomrule
\end{tabular*}
\vspace{0.5em}
\end{table}

\subsection{Rigid NVS}
\label{results:rigid}

\textit{Methods.} NeRF\cite{mildenhall2021nerf} and 3D Gaussian Splatting (3DGS)~\cite{kerbl20233d} represent current SOTA for rigid NVS: NeRF employs implicit volumetric rendering through coordinate-based MLPs, while 3DGS uses explicit anisotropic Gaussians with differentiable rasterization. We evaluate with and without depth supervision from ~\cite{yang2024depth,wen2025foundationstereo} to test if geometric priors from surgical imagery improve reconstruction quality. All experiments were conducted on processed~\cite{schoenberger2016sfm} test sequences.

\begin{figure}
\centering

\begin{minipage}{0.53\columnwidth}
    \centering
    \includegraphics[width=\linewidth]{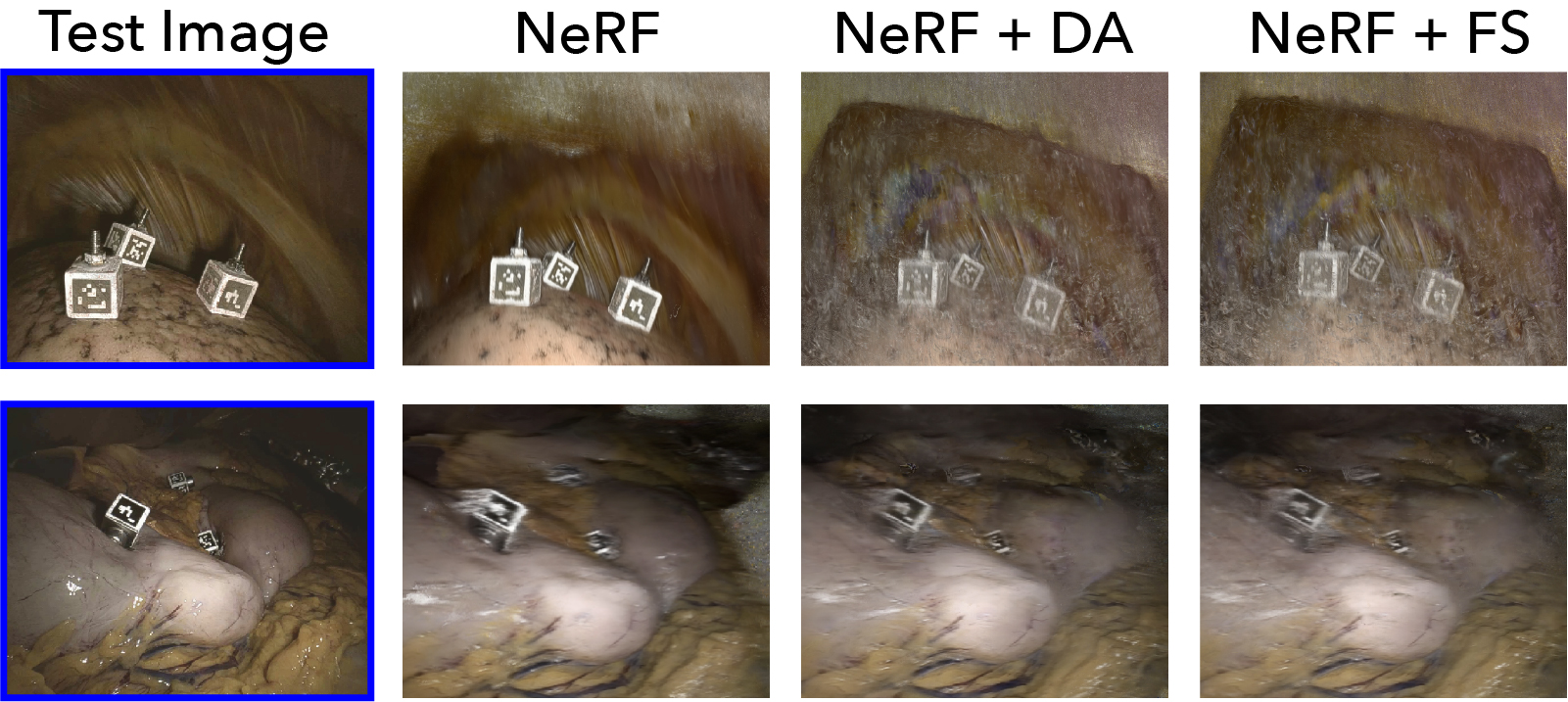}
\end{minipage}
\hfill
\begin{minipage}{0.33\columnwidth}
    \centering
    \includegraphics[width=\linewidth]{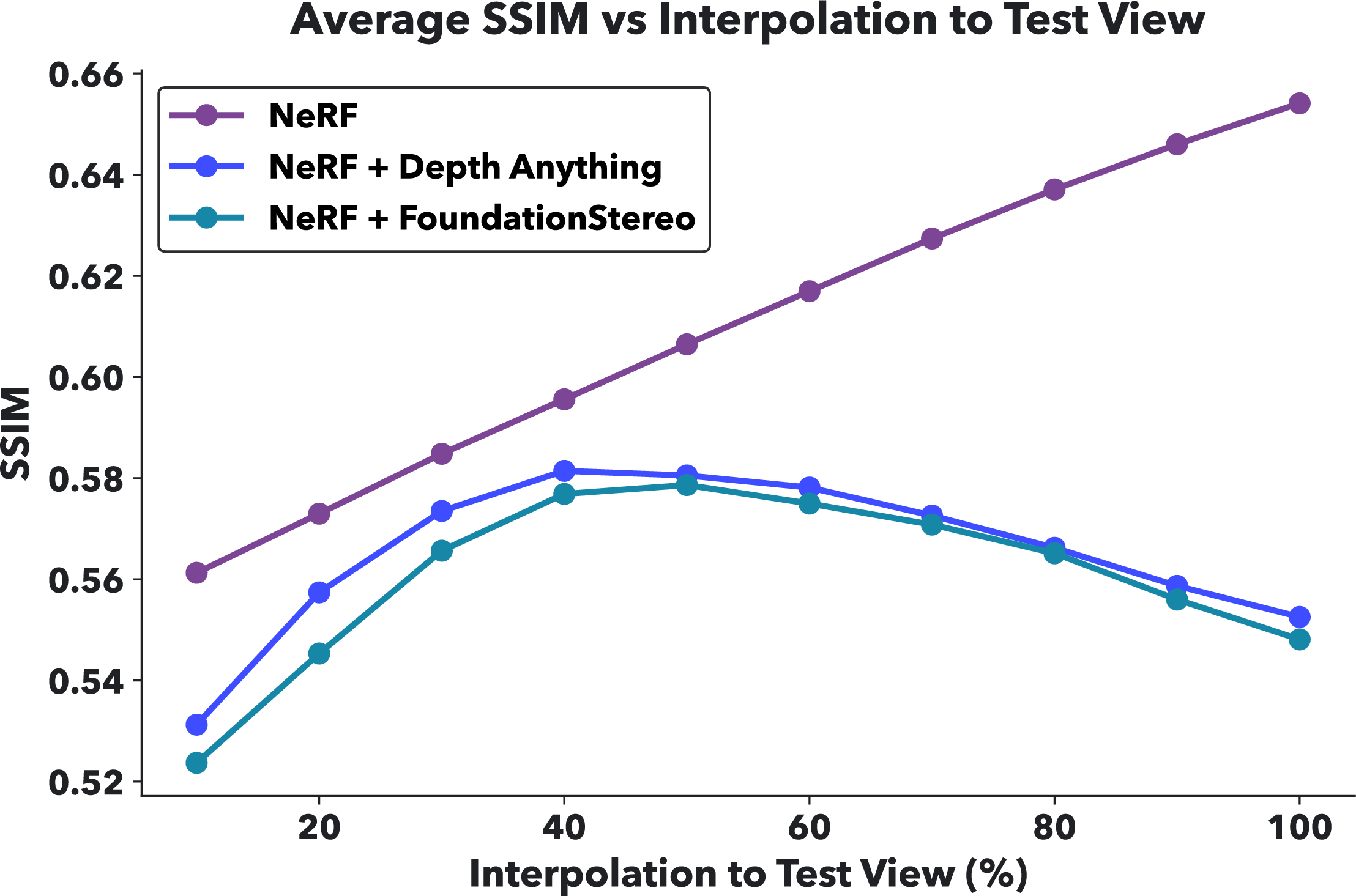}
\end{minipage}

\caption{
\textit{Left:} Rendered test views show ghost-like artifacts in depth-supervised variants. 
\textit{Right:} SSIM computed against the test held-out image during camera interpolation 
from a training (0\%) to test (100\%) viewpoint. NeRF (purple) improves monotonically, 
while depth-supervised variants degrade despite geometric convergence, suggesting 
transparent density surfaces rather than opaque geometry.
}
\label{fig:nerf_exps}
\end{figure}

\noindent\textit{Results.} Depth supervision produces opposite trends across representations: NeRF degrades, while 3DGS improves (\cref{tab:nvs-benchmark}). For NeRF, depth supervision leads to ghost-like, semi-transparent artifacts and lower fidelity at the test viewpoint (\cref{fig:nerf_exps}). To analyze this behavior, we interpolate from a train camera toward a held-out test camera in 10\% increments and compute SSIM between the rendered images and the test held-out view. In the expected case, SSIM should increase monotonically as the rendering approaches the test camera pose, which is observed for vanilla NeRF. However, dep-sup NeRF exhibits non-monotonic SSIM trends, where geometric convergence does not yield better photometric values. Prior work~\cite{turki2024hybridnerf} suggests NeRF may “cheat” by modeling semi-transparent volumes to approx. reflections; dep. sup. may accentuate this behavior, producing artifacts that worsen as views diverge from the train distribution.


\subsection{Deformable NVS}


\textit{Methods.} Live surgical scenes exhibit full field-of-view non-rigid deformation from respiratory motion, cardiac pulsation, peristalsis, and tool-tissue contact. We benchmark six deformable Gaussian Splatting variants. \textit{Offline methods} optimize over complete sequences: D-3DGS~\cite{yang2024deformable}, 4DGS (Wu et al.)~\cite{wu20244d}, 4DGS (Yang et al.)~\cite{yang2023real}, and Endo-4DGS~\cite{huang2024endo}. \textit{Online methods} perform incremental optimization during acquisition: Hayoz et al.~\cite{hayoz2024online} and DynOmo~\cite{seidenschwarz2025dynomo}. We evaluate Hayoz with both its original RAFT~\cite{teed2020raft} depth prior and FoundationStereo~\cite{wen2025foundationstereo}; DynOmo with both its original Depth Anything~\cite{yang2024depth} and FoundationStereo.

\begin{figure*}
\centering
\includegraphics[width=0.95\textwidth]{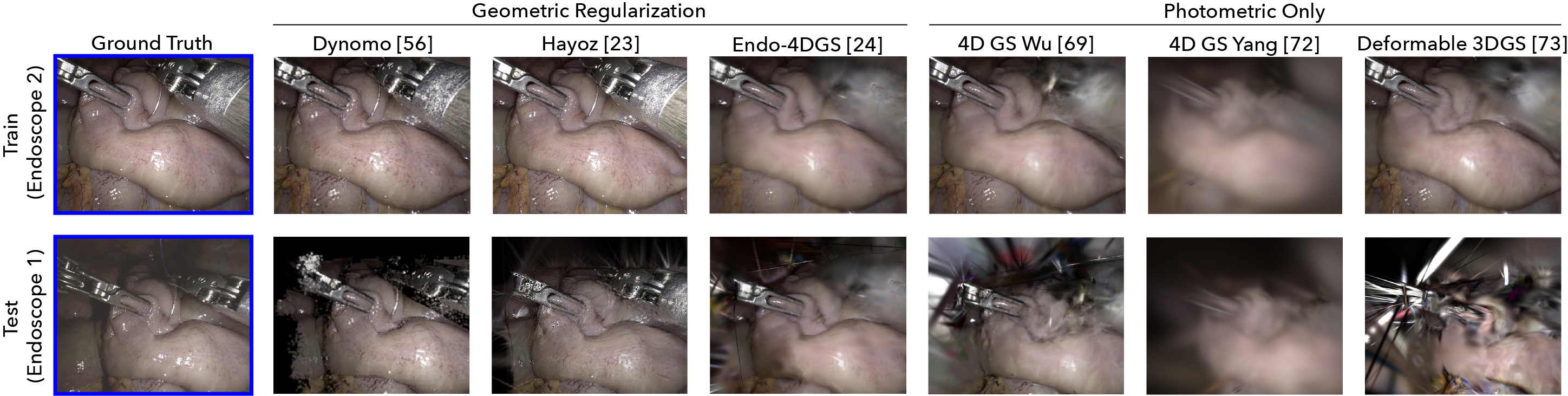}
\caption{Qualitative comparison of the Deformable NVS methods.}
\label{fig:dnvs}
\end{figure*}

\noindent\textit{Results.} Performance correlates directly with geometric regularization strength (\cref{tab:nvs-benchmark}, ~\cref{fig:dnvs}). Methods relying solely on photometric supervision fail under scenarios with constrained viewpoints, where photometric cues alone cannot disambiguate geometric structure. D-3DGS achieves low performance using only L1+D-SSIM photometric loss, allowing unconstrained per-Gaussian warping that produces geometrically inconsistent reconstructions across independent viewpoints. The 4DGS variants add temporal regularization through total variation constraints on hexplane/temporal features but lack explicit spatial coherence, achieving marginal improvements. Methods incorporating explicit 3D geometric constraints demonstrate substantial gains. Endo-4DGS achieves better quality through depth supervision, surface normal consistency, and confidence weighting—directly enforcing geometric plausibility. Hayoz applies local rigidity to sparse anchor points, providing modest spatial coherence. DynOmo achieves superior performance, applying dense 3D constraints to all Gaussians: local rigidity in position and rotation, isometry for long-term distance preservation, feature-similarity-based neighbor selection, and instance-aware grouping.

The train-on-one, test-on-another protocol measures geometric fidelity instead of photometric overfitting. Since methods without 3D constraints exhibit quality collapse at test views (\cref{fig:dnvs}), this suggests that, under this benchmark, these methods rely more on view-dependent radiance approximations than multi-view consistent geometry. DynOmo's dense spatial regularization and foundation model features maintain coherence across the $\approx$2.6cm and $\approx$18.7° viewpoint shift, achieving near-photorealistic quality, but at substantial computational cost: ~16 seconds/frame versus Endo-4DGS's ~1 second/frame in training time. This $16{\times}$ slowdown exposes a critical open problem: current 3D regularization strategies require expensive neighborhood computation, multi-constraint optimization, and dense feature matching, a computational barrier that remains prohibitive for real-time clinical deployment.

\subsection{Pose Estimation \& Feature Matching}

\begin{figure}[t]
    \centering
    \includegraphics[width=0.85\columnwidth]{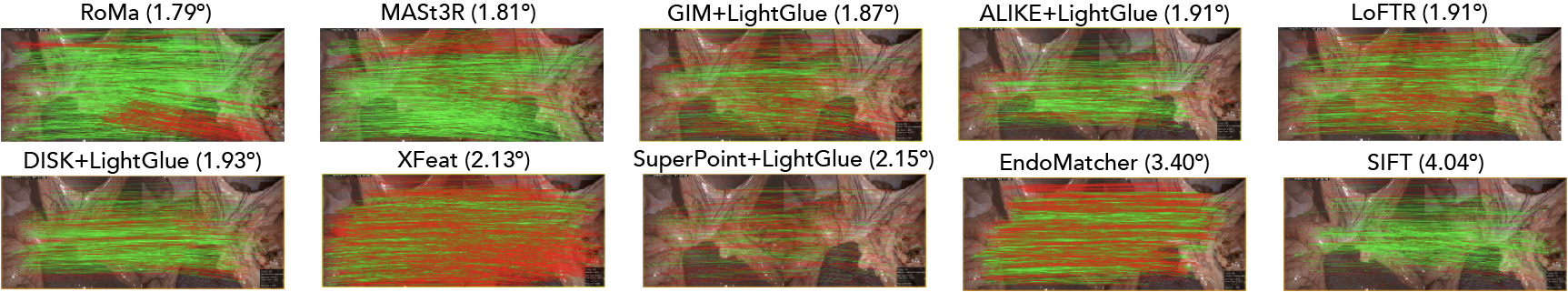}
    \caption{Qualitative comparison of image matching methods showing correspondence quality. Green lines indicate inlier matches; red lines indicate outliers. The angle next to the method name is the average rotational error.}
    \label{fig:pose_img}
\end{figure} 

\textit{Methods.} We evaluate 11 pose estimation and feature matching methods on 30 calibrated stereo endoscopy sequences with the reference camera poses. Methods include handcrafted features (SIFT~\cite{lowe2004distinctive}), learned sparse matchers (SuperPoint+LightGlue~\cite{detone2018superpoint, lindenberger2023lightglue}, DISK+LightGlue~\cite{tyszkiewicz2020disk}, ALIKE+LightGlue~\cite{zhao2023aliked}, XFeat~\cite{potje2024xfeat}), dense matchers leveraging foundation models (GIM+LightGlue~\cite{shen2024gim}, LoFTR~\cite{sun2021loftr}, EndoMatcher~\cite{yang2025endomatcher}, RoMa~\cite{edstedt2024roma}, MASt3R~\cite{leroy2024mast3r}), and a metric relative pose estimator (MicKey~\cite{barroso2024mickey}). We compute rotation error and inlier rate across all 30 sequences to measure generalization across diverse surgical scenes, anatomies, and imaging conditions. For the 23 sequences featuring tool-tissue deformations with static cameras, where scene motion occurs without camera movement, we additionally compute temporal consistency (standard deviation of rotation errors), which measures frame-to-frame stability critical for tracking and SLAM applications.
\noindent\textit{Results.} Dense foundation model methods achieve best absolute accuracy: RoMa produces 3905 inliers with 78.1\% inlier rate and 1.79° rotation error; MASt3R yields 1676 inliers with 70.3\% rate and $1.81\degree$ error, providing $2{-}50{\times}$ more correspondences than sparse methods. However, learned sparse methods achieve superior temporal consistency: GIM+LightGlue demonstrates $0.41\degree$ temporal drift versus $1.38{-}2.00\degree$ for dense methods, making it ideal for real-time SLAM despite slightly lower absolute accuracy. This temporal stability likely arises from GIM's label propagation across video frames during training, which enforces match coherence over time~\cite{shen2024gim}. MicKey~\cite{barroso2024mickey}, a metric relative pose estimator trained on outdoor scenes, reasonably failed to produce valid correspondences across all surgical sequences, presenting an opportunity for future work to develop efficient metric pose estimators robust to deformable environments. SIFT's poor performance confirms that learned methods provide better accuracy with lower variance compared to handcrafted features. These results demonstrate that models trained on large-scale diverse data transfer effectively to surgical domains, and that the accuracy-stability tradeoff between dense and sparse methods persists even in constrained imaging conditions.

\begin{table}[t]
\centering
\caption{Pose estimation and feature matching benchmarking results. Temporal consistency measures rot. drift across seqs. with tool-tissue deformation and static cameras.}
\label{tab:pose-feature-benchmark}
\scriptsize
\setlength{\tabcolsep}{3pt}
\begin{tabular*}{\columnwidth}{@{\extracolsep{\fill}}lcccc}
\toprule
Method & \makecell{Rot. Err.(°)$\downarrow$} & \makecell{Temporal Drift(°)$\downarrow$} & \makecell{Inlier Rate (\%)$\uparrow$} & \makecell{Avg Inliers$\uparrow$} \\
\midrule
RoMa~\cite{edstedt2024roma} & 1.79±1.41 & 1.41 & 78.1 & 3905 \\
MASt3R~\cite{leroy2024mast3r} & 1.81±1.38 & 1.38 & 70.3 & 1676 \\
GIM+LG~\cite{shen2024gim,lindenberger2023lightglue} & 1.87±1.48 & 0.41 & 43.1 & 264 \\
ALIKE+LG~\cite{zhao2023aliked,lindenberger2023lightglue} & 1.91±1.52 & 0.64 & 68.2 & 211 \\
LoFTR~\cite{sun2021loftr} & 1.91±1.40 & 0.43 & 47.2 & 513 \\
DISK+LG~\cite{tyszkiewicz2020disk,lindenberger2023lightglue} & 1.93±1.52 & 0.59 & 55.2 & 446 \\
XFeat~\cite{potje2024xfeat} & 2.13±1.47 & 0.77 & 25.9 & 854 \\
SP+LG~\cite{detone2018superpoint,lindenberger2023lightglue} & 2.15±1.41 & 0.70 & 34.2 & 74 \\
EndoMatcher~\cite{yang2025endomatcher} & 3.40±1.91 & 2.11 & 39.9 & 1255 \\
SIFT~\cite{lowe2004distinctive} & 4.04±14.45 & 5.44 & 52.9 & 91 \\
\bottomrule
\end{tabular*}
\vspace{0.5em}
\end{table}

\subsection{Monocular Depth Estimation}

\begin{figure}[t]
    \centering
    \includegraphics[width=0.9\columnwidth]{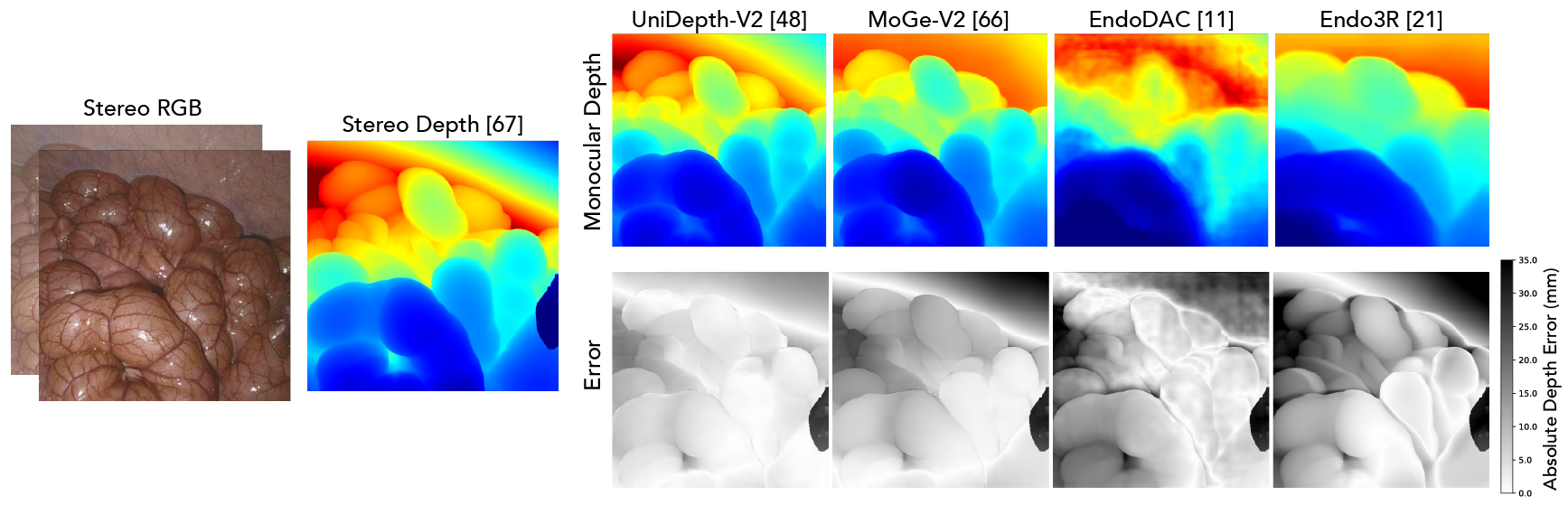}
    \caption{Qualitative comparison of mono. depth estimation compared to \textit{inferred} stereo depth~\cite{wen2025foundationstereo}. Error maps show abs. depth error after median scaling of the predictions.}
    \label{fig:depth_vis}
\end{figure} 

\textit{Methods.}
We evaluate monocular depth prediction on calibrated stereo endoscopy sequences using reference depth obtained from a SOTA stereo prior~\cite{wen2025foundationstereo}.  Given a left image, each model predicts dense depth $\hat{d}_i$, which we compare to the stereo reference $d_i$ reprojected into the same frame. We benchmark four representative models without sequence-specific fine-tuning: two domain-specific surgical models, Endo3R~\cite{guo2025endo3r} and EndoDAC~\cite{cui2024endodac}, and two recent metric-scale SOTA monocular depth models, UniDepth-V2~\cite{piccinelli2025unidepthv2} and MoGe-V2~\cite{wang2025moge}. 
Evaluation is performed on the same 30 test sequences from ex-vivo, in-vivo, and postmortem scenes (over $6\times10^8$ valid pixels). We report standard depth metrics: absolute relative error (Abs Rel), root mean squared error (RMSE), log RMSE, and threshold accuracies $\delta_1$, $\delta_2$, $\delta_3$, where 
$\max(\hat{d}_i/d_i, d_i/\hat{d}_i) < 1.25^k$, $k\!\in\!\{1,2,3\}$. Since monocular depth is inherently ambiguous in global scale, we report results \emph{with median scaling} (single optimal scale per sequence). For two models that explicitly claim metric-scale prediction~\cite{piccinelli2025unidepthv2,wang2025moge}, we also report \emph{without scaling}.

\textit{Results.} With median scaling, the two general-purpose, metric-scale monocular depth models~\cite{piccinelli2025unidepthv2,wang2025moge}) outperform the surgical-domain methods~\cite{cui2024endodac,guo2025endo3r}(\cref{tab:depth_with_scaling}, \cref{fig:depth_vis}). This holds despite EndoDAC being initialized via~\cite{yang2024depth} and subsequently fine-tuned on surgical data, while Endo3R is trained only on surgical imagery. In the no-scaling setting, the two models that claim metric scale exhibit near-zero $\delta$ accuracies and large errors. This indicates a substantial global scale mismatch and reinforces that explicit scale calibration is necessary in surgical deployment. 


\begin{table}[t]
\centering
\caption{Monocular depth prediction compared to \textit{inferred} stereo depth~\cite{wen2025foundationstereo}. 
Methods that report metric-scale predictions~\cite{piccinelli2025unidepthv2,wang2025moge} are evaluated without median scaling.}
\label{tab:depth_with_scaling}
\scriptsize
\setlength{\tabcolsep}{2pt}
\begin{tabular*}{\columnwidth}{@{\extracolsep{\fill}}lccccccc}
\toprule
Method 
& Med. Scaled
& Abs Rel$\downarrow$ 
& RMSE$\downarrow$ 
& RMSE$_{\log}$$\downarrow$ 
& $\delta_1\uparrow$ 
& $\delta_2\uparrow$ 
& $\delta_3\uparrow$ \\
\midrule

UniDepth-V2~\cite{piccinelli2025unidepthv2} 
& Yes
& 0.081
& 0.012 
& 0.111 
& 0.936
& 0.992
& 0.998 \\

MoGe-V2~\cite{wang2025moge} 
& Yes
& 0.109 
& 0.014 
& 0.140 
& 0.889 
& 0.989 
& 0.998 \\

EndoDAC~\cite{cui2024endodac} 
& Yes
& 0.119 
& 0.017 
& 0.164 
& 0.854 
& 0.968 
& 0.994 \\

Endo3R~\cite{guo2025endo3r} 
& Yes
& 0.187 
& 0.023 
& 0.236 
& 0.704 
& 0.927 
& 0.972 \\

\midrule

UniDepth-V2~\cite{piccinelli2025unidepthv2} 
& No
& 11.573
& 1.098 
& 2.465 
& 0.000
& 0.000
& 0.000 \\

MoGe-V2~\cite{wang2025moge} 
& No
& 14.556 
& 1.329 
& 2.683 
& 0.000 
& 0.000 
& 0.000 \\

\bottomrule
\end{tabular*}
\vspace{0.3em}
\end{table}

\noindent\textbf{Limitations.} \textit{Ex vivo} sequences use hand-guided (non-robotic) endoscope motion, resulting in less smooth trajectories than robotic manipulation. For dynamic camera sequences where ArUco calibration markers remain visible throughout, mask-based preprocessing is used to exclude markers~\cite{li2025diffueraser}. Viewpoint diversity is lower than classic multi-view datasets because surgical acquisition is constrained by trocar placement, visibility, and safe camera motion. More subject and anatomical diversity is important future work.

\section{Conclusion}
\label{sec:conclusion}

iMED exposes limitations in 3D perception under constrained, deformable surgical environments through synchronized recordings from two independent stereo endoscopes with overlapping fields of view. This enables evaluation from physically distinct viewpoints, revealing whether methods recover consistent geometry or mainly interpolate appearance along a single trajectory. Across rigid, deformable, and online novel view synthesis, pose estimation/feature matching, and monocular depth estimation, many methods degrade under narrow-baseline motion, specular tissue, full-field deformation, and topology changes. Surgical scenes therefore act as a stress test for assumptions often hidden by standard same-trajectory evaluations.

Across tasks, broad multi-domain vision models often transfer more robustly to unseen surgical domains than surgical-only models or aggressively fine-tuned methods, suggesting that narrow training can discard invariances needed for cross-device and cross-site generalization. We hypothesize that robust surgical 3D perception will require multi-domain training that incorporates surgical data without sacrificing general visual priors. iMED provides infrastructure for this direction, with diverse anatomical conditions and a benchmark that rewards genuine geometric generalization. Advances under these constraints are likely to transfer to robotics, AR/VR, and embodied systems operating in dynamic, sensor-limited environments.
\noindent\textbf{Acknowledgements.} This work was supported in part by Intuitive Surgical and by the Engineering and Physical Sciences Research Council (EPSRC) through the OASIS programme under grant UKRI145.


%
%
\bibliographystyle{splncs04}
\bibliography{main}

\clearpage
\setcounter{page}{1}
\section*{Supplementary Material}

\label{sec:appendix}

\section{Dataset Curation: Detailed Methodology}
\label{app:dataset_curation}

This supplementary provides technical details for data acquisition, calibration, and validation procedures described in Section~\ref{sec:dataset_curation}.

\subsection{Hardware Synchronization \& Compression}
\label{app:hardware_sync}

Four HDMI streams (left and right cameras from each endoscope) were captured at 1280×1024 resolution and 60 Hz using a four channel HD capture card (Pro Capture Quad HDMI, Magewell) interfaced via Video4Linux2 API. Synchronized capture employed four parallel processes—two for frame grabbing and two for disk writing with sub millisecond-precision timestamps. Synchronization accuracy across $\approx$170K frame pairs achieved mean timing difference of 3.36 ms (SD = 3.90 ms, median = 3.00 ms), with 99th percentile at 17 ms. Only 1.13\% of frames exceeded the single-frame threshold (16 ms), attributable to independent camera shutters, clock drift, HDMI transmission latency, and disk scheduling; these can be filtered using provided timestamps. 

Given the substantial size of our dataset, extracting all frames would result in well over 1 TB of storage—presenting significant storage and open sourcing challenges. We compress all sequences using H.264 CRF 18 encoding. We strongly recommend direct video processing using NVIDIA's hardware decoder (NVDEC) with GPU memory transfer for training and inference pipelines, which bypasses CPU-based frame extraction and reduces I/O overhead. For workflows requiring CPU-only processing (\eg, COLMAP-based reconstruction) or researchers with limited storage capacity, we provide task-specific dataset train/test subset recommendations in Tab~\ref{tab:dataset_usage}.

\subsection{Calibration Cube Design}
\label{app:cube_design}

Our calibration cube design evolved through multiple iterations to address detection challenges under surgical imaging conditions (Figure~\ref{fig:aruco_combined} in main paper). Initial stainless steel cubes with "\texttt{ARUCO\_4x4\_DICT}" markers suffered from specular reflections under point-source illumination and limited error correction from the smaller dictionary. These design limitations contributed to the high baseline trajectory errors observed in early ex vivo sequences (Table~\ref{tab:trajectory-comparison-appendix}, 362-550 pixel reprojection error). We converged on 8mm wooden cubes painted white with 5.7mm "\texttt{ARUCO\_MIP\_36h12}" markers~\cite{garrido2016generation}, providing 12-bit hamming distance error correction. Markers were laser-etched using 3W UV (355nm, IPG Photonics) and sealed with clear acrylic spray (Krylon Matte Finish Clear Coating) to prevent staining. Each cube included a screw or inverted screw for robotic manipulation and trocar retrieval. The painted wooden substrate eliminates specular artifacts while maintaining high-contrast detection, as evidenced by the dramatic improvement in trajectory accuracy for sequences using the improved design.

\subsection{Calibration Protocols}
\label{app:calibration_protocol}

Two distinct acquisition protocols based on ArUcos were used, filterable via metadata flags:
\noindent(1) \textit{Static Camera Post-Calibration Sequences:} ($\approx$40\% of sequences)
For tissue manipulation tasks, we first perform a calibration phase where ArUco cubes are positioned throughout the surgical field and manipulated with robotic instruments to densely sample the working volume. After acquiring sufficient observations for robust camera-to-camera calibration, the cubes are removed and tissue manipulation proceeds with fixed camera positions. This protocol provides reference poses without visual interference during task execution. To quantify calibration stability, we conducted pre- and post-task recalibration experiments across 13 tissue manipulation sequences. Comparing camera-to-camera transformations before and after task completion revealed mean rotational drift of 1.97° (median: 1.93°) and mean translational drift of 3.3mm (median: 2.2mm). These drift values are consistent with minor mechanical settling of the passive articulated arm during manipulation. While all sequences are pre-processed with computed poses, raw calibration and marker tracking seqs are included for users who wish to perform custom calibration.

\noindent(2) \textit{Dynamic Camera Sequences:} ($\approx$40\% of sequences)
For sequences involving endoscope movement, ArUco markers remain visible throughout to enable continuous pose estimation. These sequences are labeled with a "\texttt{*\_camera\_movement}" flag in the metadata, allowing task-specific filtering. For dynamic-camera sequences where markers remain visible for continuous pose estimation, released frames include ArUco masks and inpainted versions generated using DiffuEraser~\cite{li2025diffueraser}; static-camera benchmark sequences remove markers before task execution.

\begin{figure}[t]
    \centering
    \includegraphics[width=0.7\columnwidth]{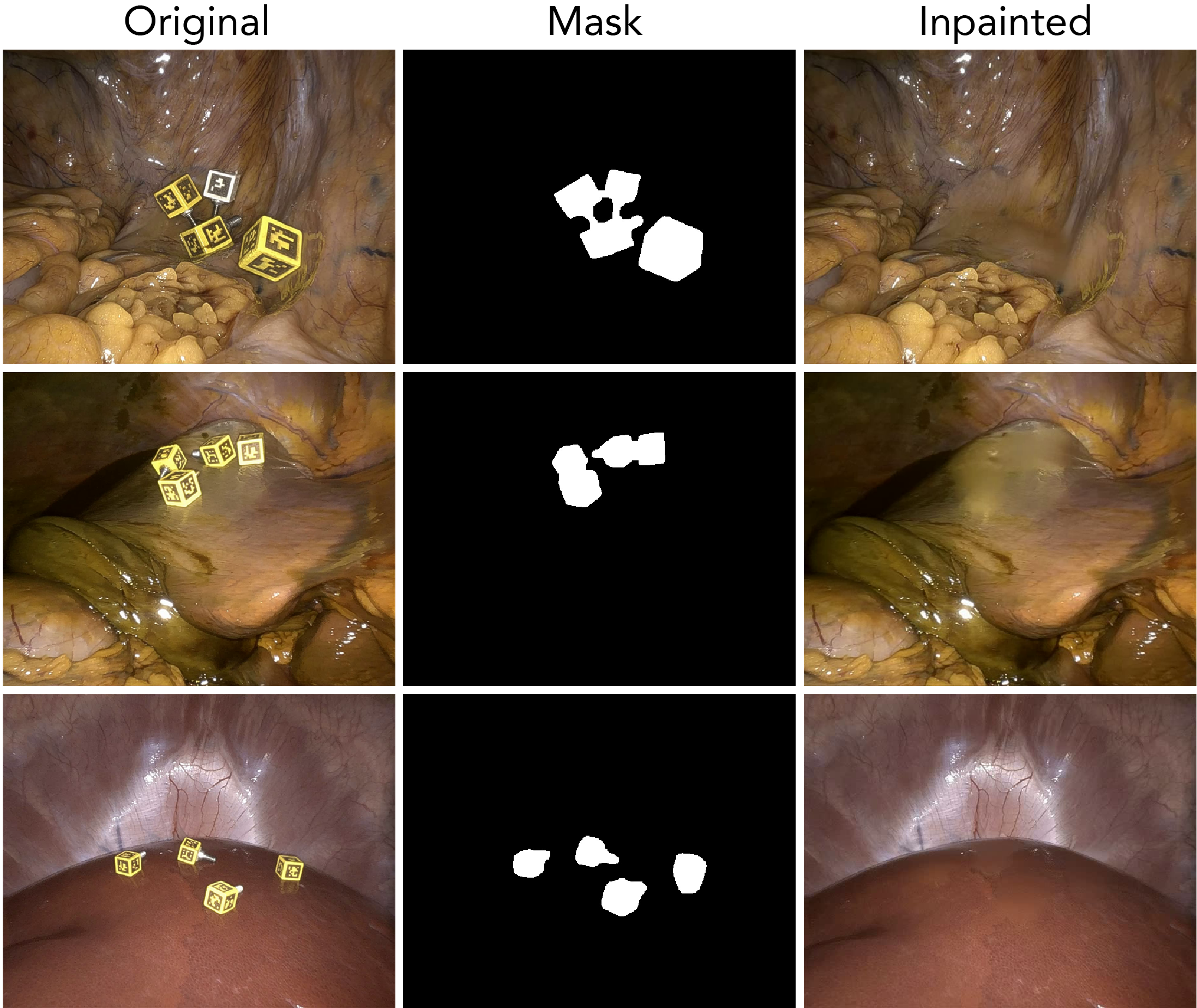}
    \caption{
    Example ArUco marker removal for dynamic-camera sequences. 
    When markers remain visible for continuous pose estimation, iMED releases ArUco masks~\cite{carion2025sam} and inpainted frames generated with DiffuEraser~\cite{li2025diffueraser}. 
    }
    \label{fig:aruco_inpainting}
\end{figure}

\subsection{Endoscope Photometric Matching}
\label{app:photometric_matching}

Since only Endoscope 2's integrated light source is activated during acquisition (to maintain representative MIS lighting conditions), the two endoscopes observe the scene under different illumination characteristics—Endoscope 1 receives illumination from Endoscope 2's displaced light source while Endoscope 2 has co-located illumination. This creates photometric discrepancies between the two endoscope views. We therefore normalize their photometric appearance to enable consistent multi-view processing. 

We evaluated four normalization strategies on nine sequences (20 frames each), applied independently to RGB channels: \textit{M0 (Baseline)} with no adjustment; \textit{M1 (Brightness Only)} scaling by grayscale mean intensity ratio; \textit{M2 (Frame-to-Frame)} aligning per-channel brightness and contrast using frame statistics; \textit{M3 (Frame-to-Sequence)} as M2 but computing target statistics once per sequence; and \textit{M4 (Sequence-to-Sequence)} computing both source and target statistics across full sequences. For M2--M4, we evaluated masking specular pixels at various thresholds. Detailed results are in Supplementary~\ref{tab:endoscope-matching-appendix}. We selected M4 to ensure robustness against per-frame artifacts (smoke, occlusions, etc.) by applying a consistent global transformation computed from entire sequence statistics rather than adapting to potentially anomalous individual frames.

\subsection{Trajectory Quality Validation}
\label{app:trajectory_validation}

To validate our trajectory pipeline design choices, we evaluated four trajectory estimation methods on 27 sequences (9 per session type): \textit{Inlier Averaging} (RANSAC baseline), \textit{Reprojection Error Optimization} (adding Levenberg-Marquardt refinement), \textit{Outlier Rejection} (adding velocity-based outlier detection), and \textit{Temporal Smoothing} (combining outlier rejection with Savitzky-Golay filtering~\cite{orfanidis1995introduction}).

\subsubsection{Evaluation Methodology}

We assessed trajectory quality using 2D reprojection error (detecting ArUco markers in both cameras, applying estimated transformations, measuring pixel-space error) and 3D geometric consistency (computing RMSE and Chamfer distance between transformed point clouds from FoundationStereo depth maps). We sampled 20 frames per sequence and convert 2D errors to physical errors at 5cm working distance using $\text{Error}_{3D} = \text{Error}_{2D} \times \text{depth} / f_x$. Full results in Supplementary~\ref{tab:trajectory-comparison-appendix}.

\subsubsection{Results and Analysis}

\textit{Ex vivo sessions} show highest baseline errors (362-550 pixels, 1.78-2.70cm at 5cm depth) but benefit dramatically from temporal smoothing (38.33 pixels, 0.19cm). Higher baseline errors stem from earlier collection phases using stainless steel cubes with 4×4 ArUco markers (~\cref{fig:aruco_combined}), which suffered from specular reflections and limited error correction. Subsequent sequences using wooden cubes with "\texttt{ARUCO\_MIP\_36h12}" markers achieved substantially improved detection and lower errors. \textit{In vivo postmortem sessions} achieve decent baseline performance (17-19 pixels, 0.08-0.10cm) with some benefit from temporal smoothing, indicating the RANSAC+optimization pipeline adequately handles these scenes where marker detection is less erroneous. \textit{In vivo live sessions} demonstrate best baseline performance (12-15 pixels, 0.06-0.07cm). Temporal smoothing provides modest improvement (11.87 pixels, 0.058cm). 3D metrics generally corroborate 2D findings, showing ex-vivo improvements from 1.5-1.8cm to 1.4-1.7cm and consistent 0.5-0.7cm performance for in-vivo sequences. However, note that 3D metrics depend on stereo depth quality and introduce additional noise beyond pose estimation error.

\begin{table}[t]
\centering
\caption{Surgical instrument taxonomy and sequence frequency across the dataset (20 instrument types, 340 sequences total).}
\label{tab:tool_taxonomy}
\scriptsize
\setlength{\tabcolsep}{2pt}
\begin{tabular}{lc}
\toprule
\textbf{Tool Name} & \textbf{Num Sequences} \\
\midrule
Intuitive Surgical Da Vinci Tip-Up Fenestrated Grasper & 61 \\
Intuitive Surgical Da Vinci ProGrasp\textsuperscript{tm} Forceps & 30 \\
Intuitive Surgical Da Vinci Long Bipolar Grasper & 22 \\
Intuitive Surgical Da Vinci Monopolar Curved Scissors & 18 \\
Ethicon Endopath XCEL Universal Sleeve Assist Port CB5ST & 18 \\
Intuitive Surgical Da Vinci Mega SutureCut\textsuperscript{tm} Needle Driver & 16 \\
Intuitive Surgical Da Vinci Cadiere Forceps & 10 \\
Intuitive Surgical Da Vinci Maryland Bipolar Forceps & 9 \\
Intuitive Surgical Da Vinci Fenestrated Bipolar Grasper & 10 \\
Ethicon 5DCS Curved Scissors & 9 \\
Intuitive Surgical Da Vinci Curved Bipolar Dissector & 7 \\
Intuitive Surgical Da Vinci Force Bipolar Grasper & 6 \\
Intuitive Surgical Da Vinci Large Needle Driver & 5 \\
Intuitive Surgical Da Vinci Resano Forceps & 5 \\
Intuitive Surgical Da Vinci Permanent Cautery Hook & 4 \\
Ethicon 5DCD Curved Dissector & 4 \\
Aven Tools Chain Nose Pliers & 4 \\
Stryker StrykeFlow II pump & 3 \\
Intuitive Surgical Da Vinci Large Clip Applier & 2 \\
Ethicon 5BB Babcock & 2 \\
\bottomrule
\end{tabular}
\end{table}

\begin{figure}[t]
\centering
\includegraphics[width=0.8\columnwidth]{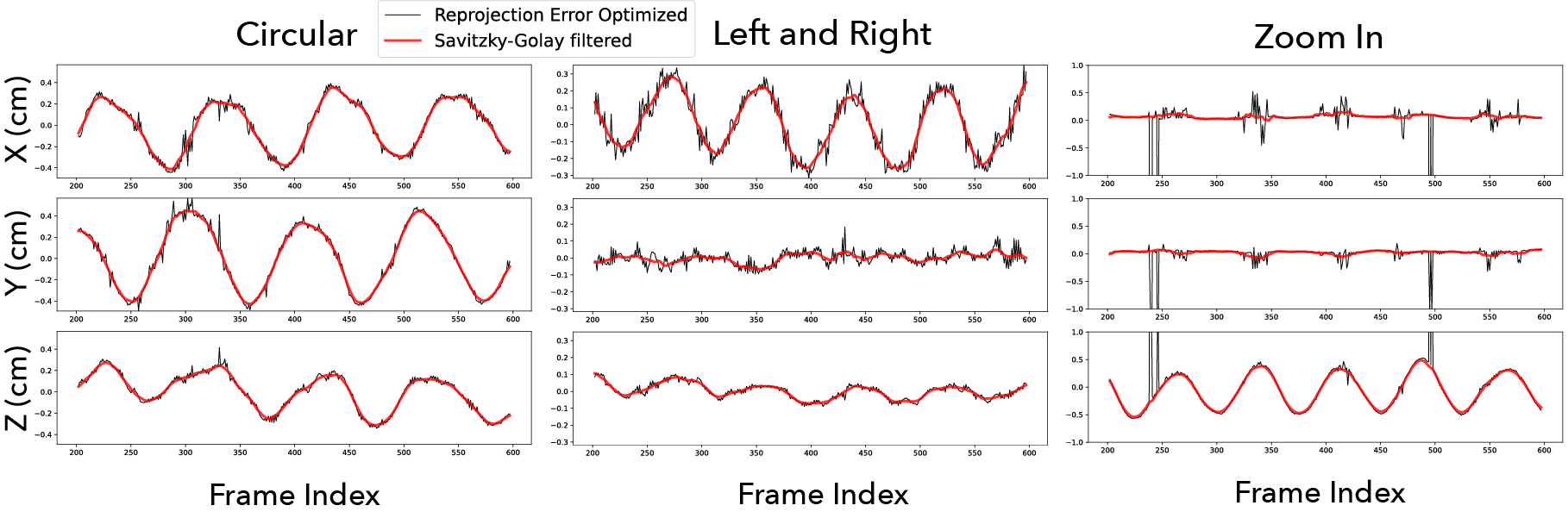}
\caption{Three trajectory examples showing camera movements in iMED - Circular (left), Left and Right(middle), zoom in (right) camera movements. They are decomposed to their X, Y, and Z (cm) components; reprojection error optimization baseline in black and the filtered trajectories in red.}
\label{fig:trajectory_examples}
\end{figure}

\begin{table}[h]
\centering
\caption{Endoscope matching normalization strategies. SSIM and PSNR measure structural and pixel-level fidelity, while temporal variation quantifies stability across frames ($\approx$ M0 baseline indicates optimal preservation of natural temporal dynamics). Methods M2-M4 evaluated with specular masking; best threshold reported. Temporal variation measures pixel stability across frames (RMS of standard deviations from 10 tracked non-specular pixels).}
\label{tab:endoscope-matching-appendix}
\small
\begin{tabular}{lccc}
\toprule
Method &SSIM$\uparrow$&PSNR$\uparrow$(dB)&\makecell{Temporal Var.\\$\approx$ M0 (6.69)}\\
\midrule
\textbf{M0}: Baseline & $0.278$ & $9.79$ & $6.69$ \\
\textbf{M1}: Brightness Only & $0.283$ & $12.10$ & $13.88$ \\
\textbf{M2}: Frame-to-Frame & $0.479$ & $15.06$ & $9.55$ \\
\textbf{M3}: Frame-to-Seq & $\mathbf{0.502}$ & $\mathbf{15.34}$ & $\mathbf{7.64}$ \\
\textbf{M4}: Seq-to-Seq & $\mathbf{0.503}$ & $\mathbf{15.34}$ & $\mathbf{7.78}$ \\
\bottomrule
\end{tabular}
\end{table}

\label{appendix:trajectory_methods}
\begin{table*}[t]
\centering
\caption{Trajectory estimation comparison across session types on 27 sequences (9 per type, 20 frames each). 2D reprojection error converted to physical error at 5cm working distance. 3D metrics from FoundationStereo point cloud alignment. Ex vivo baseline errors (362-550 pixels) stem from early acquisition with metallic cubes and 4×4 ArUco markers; temporal smoothing yields dramatic improvement. In vivo sessions achieve lower baseline errors (12-19 pixels) due to improved marker design (wooden cubes, "\texttt{ARUCP\_MIP\_36h12}") and mechanical stability.}
\label{tab:trajectory-comparison-appendix}
\scriptsize
\begin{tabular*}{\textwidth}{@{\extracolsep{\fill}}lccccc}
\toprule
& \multicolumn{3}{c}{2D Evaluation} & \multicolumn{2}{c}{3D Evaluation} \\
\cmidrule(lr){2-4} \cmidrule(lr){5-6}
Method & \makecell{Reproj. Err.\\(pixels)$\downarrow$} & \makecell{Physical Err.\\@5cm (cm)$\downarrow$} & \makecell{Markers/\\Image Pair$\uparrow$} & \makecell{RMSE(cm)$\downarrow$} & \makecell{Chamfer Dist.\\(cm)$\downarrow$} \\
\midrule
\multicolumn{6}{c}{Ex Vivo Sessions (n=9)} \\
\midrule
Inlier Averaging & 550.23 & 2.697 & 4.18 & 1.5 & 1.8 \\
Optimization & 362.69 & 1.778 & 4.18 & 1.5 & 1.8 \\
Outlier Rejection & 362.69 & 1.778 & 4.18 & 1.5 & 1.8 \\
Temporally Smooth & 38.33 & 0.188 & 4.18 & 1.4 & 1.7 \\
\midrule
\multicolumn{6}{c}{In Vivo Postmortem Sessions (n=9)} \\
\midrule
Inlier Averaging & 19.30 & 0.095 & 3.45 & 0.5 & 0.6 \\
Optimization & 17.00 & 0.083 & 3.45 & 0.5 & 0.6 \\
Outlier Rejection & 17.00 & 0.083 & 3.45 & 0.5 & 0.6 \\
Temporally Smooth & 19.89 & 0.098 & 3.45 & 0.5 & 0.6 \\
\midrule
\multicolumn{6}{c}{In Vivo Live Sessions (n=9)} \\
\midrule
Inlier Averaging & 12.04 & 0.059 & 5.34 & 0.7 & 0.7 \\
Optimization & 14.50 & 0.071 & 5.34 & 0.7 & 0.7 \\
Outlier Rejection & 14.77 & 0.072 & 5.34 & 0.7 & 0.7 \\
Temporally Smooth & 11.87 & 0.058 & 5.34 & 0.7 & 0.7 \\
\bottomrule
\end{tabular*}
\end{table*}



\begin{sidewaystable}
\centering
\caption{Task-specific filtering via dataloader API for researchers to subset sequences by research objective. *All \textit{in vivo} live sequences inherently contain respiratory motion and continuous tissue deformation. **Temporal evaluation requires sequences with sustained camera motion for trajectory consistency assessment.}
\label{tab:dataset_usage}
\scriptsize
\begin{tabular}{l p{4cm} p{4cm} p{4cm}}
\toprule
Task & Ex Vivo & In Vivo Postmortem & In Vivo Live \\
\midrule
Rigid Novel View Synthesis & 
\texttt{dataset\_loader.filter(} \newline
\texttt{~~session\_types=["ex\_vivo"],} \newline
\texttt{~~movement\_types=} \newline
\texttt{~~~~["circular\_camera\_motion",} \newline
\texttt{~~~~~"zoom\_camera\_motion",} \newline
\texttt{~~~~~"lateral\_camera\_motion"])} & 
\texttt{dataset\_loader.filter(} \newline
\texttt{~~session\_types=} \newline
\texttt{~~~~["in\_vivo\_postmortem"],} \newline
\texttt{~~movement\_types=} \newline
\texttt{~~~~["circular\_camera\_motion",} \newline
\texttt{~~~~~"zoom\_camera\_motion",} \newline
\texttt{~~~~~"lateral\_camera\_motion"])} & 
N/A \\
\midrule
Deformable Novel View Synthesis & 
\texttt{dataset\_loader.filter(} \newline
\texttt{~~session\_types=["ex\_vivo"],} \newline
\texttt{~~movement\_types=} \newline
\texttt{~~~~["off\_camera\_manipulation",} \newline
\texttt{~~~~~"tool\_manipulation"])} & 
\texttt{dataset\_loader.filter(} \newline
\texttt{~~session\_types=} \newline
\texttt{~~~~["in\_vivo\_postmortem"],} \newline
\texttt{~~movement\_types=} \newline
\texttt{~~~~["off\_camera\_manipulation",} \newline
\texttt{~~~~~"tool\_manipulation"])} & 
\texttt{dataset\_loader.filter(} \newline
\texttt{~~session\_types=} \newline
\texttt{~~~~["in\_vivo\_live"])}* \\
\midrule
\makecell[l]{Pose Estimation,\\Optical Flow Eval.,\\Monocular Depth Est.} & 
\texttt{dataset\_loader.filter(} \newline
\texttt{~~session\_types=["ex\_vivo"],} \newline
\texttt{~~aruco\_present=False)} & 
\texttt{dataset\_loader.filter(} \newline
\texttt{~~session\_types=} \newline
\texttt{~~~~["in\_vivo\_postmortem"],} \newline
\texttt{~~aruco\_present=False)} & 
\texttt{dataset\_loader.filter(} \newline
\texttt{~~session\_types=} \newline
\texttt{~~~~["in\_vivo\_live"],} \newline
\texttt{~~aruco\_present=False)} \\
\midrule
\makecell[l]{Temporal Stereo\\Pose Estimation**} & 
\texttt{dataset\_loader.filter(} \newline
\texttt{~~session\_types=["ex\_vivo"],} \newline
\texttt{~~movement\_types=} \newline
\texttt{~~~~["circular\_camera\_motion",} \newline
\texttt{~~~~~"zoom\_camera\_motion",} \newline
\texttt{~~~~~"lateral\_camera\_motion"])} & 
\texttt{dataset\_loader.filter(} \newline
\texttt{~~session\_types=} \newline
\texttt{~~~~["in\_vivo\_postmortem"],} \newline
\texttt{~~movement\_types=} \newline
\texttt{~~~~["circular\_camera\_motion",} \newline
\texttt{~~~~~"zoom\_camera\_motion",} \newline
\texttt{~~~~~"lateral\_camera\_motion"])} & 
\texttt{dataset\_loader.filter(} \newline
\texttt{~~session\_types=} \newline
\texttt{~~~~["in\_vivo\_live"],} \newline
\texttt{~~movement\_types=} \newline
\texttt{~~~~["circular\_camera\_motion",} \newline
\texttt{~~~~~"zoom\_camera\_motion",} \newline
\texttt{~~~~~"lateral\_camera\_motion"])} \\
\bottomrule
\end{tabular}
\end{sidewaystable}
\end{document}